\PassOptionsToPackage{dvipsnames}{xcolor}
\documentclass[preprint,12pt,a4paper]{elsarticle}

\usepackage[utf8]{inputenc}
\usepackage{arydshln}
\usepackage{natbib}
\setcitestyle{square, numbers,sort}
\usepackage{amssymb}
\usepackage{amsmath}
\usepackage{lineno}
\usepackage[bookmarks=true,bookmarksnumbered=true]{hyperref}
\usepackage{graphicx}
\usepackage{tikz}
\usepackage{forest}
\usepackage{stanli}
\usetikzlibrary{positioning,shapes,shadows,arrows ,automata,arrows.meta,decorations, decorations.text,calligraphy}
\usepackage{pgfplots}
\pgfplotsset{compat=1.18}
\usepackage{booktabs}
\usepackage{varwidth}
\usepackage{listings}
\usepackage{ragged2e}
\usepackage{algorithm}
\usepackage[noend]{algpseudocode}
\usepackage{dsfont}
\usepackage{todonotes}
\usepackage{color,soul}
\usepackage{enumitem}
\usepackage{amsthm}
\usepackage{xcolor}
\usepackage{subfig}
\if@symbold\else\fi

\newcommand{\eqnref}[1]{Eq.~(\ref{#1})}
\newcommand{\figref}[1]{Figure~\ref{#1}}
\newcommand{\tabref}[1]{Table~\ref{#1}}

\newcommand{\meta}{\mathrm{met}}
\newcommand{\cat}{\mathrm{cat}}

\newcommand{\quant}{\mathrm{qnt}}

\newcommand{\neutral}{\mathrm{neu}}
\newcommand{\decreed}{\mathrm{dec}}
\newcommand{\acting}{\mathrm{inc}}

\newcommand{\tstar}[5]{
\pgfmathsetmacro{\starangle}{360/#3}
\draw[#5] (#4:#1)
\foreach \x in {1,...,#3}
{ -- (#4+\x*\starangle-\starangle/2:#2) -- (#4+\x*\starangle:#1)
}
-- cycle;
}

\makeatletter
\newcommand*\getangle[2]{
\pgfextractx{\pgf@xa}{\pgfpointanchor{#2}{center}}
\pgfextracty{\pgf@ya}{\pgfpointanchor{#2}{center}}
\pgfmathsetmacro{#1}{atan2(\pgf@xa,\pgf@ya)}
}
\makeatother
\usepackage []{csquotes}

\MakeOuterQuote{"}

\journal{Advances in Engineering Software}

\begin{document}

\newtheorem{assumption}{Assumption}
\newtheorem{theorem}{Theorem}
\newtheorem{corollary}{Corollary}
\newtheorem{lemma}{Lemma}
\newtheorem{remark}{Remark}

\lstset{
language=Python,
morekeywords={smt}
emphstyle=\color{red}, 
basicstyle=\ttfamily \small,
frame=single,
showspaces=false,
showstringspaces=false,
keywordstyle=\ttfamily \color{blue},
commentstyle=\color{green!30!red!70}\ttfamily
}

\begin{frontmatter}
\title{\texttt{SMT 2.0}: A Surrogate Modeling Toolbox with a focus on Hierarchical and Mixed Variables Gaussian Processes}

\author[label3,label7]{Paul Saves}
\ead{paul.saves@onera.fr} 

\author[label3]{R\'emi Lafage}
\ead{remi.lafage@onera.fr}

\author[label3]{Nathalie Bartoli}
\ead{nathalie.bartoli@onera.fr}

\author[label6]{Youssef Diouane}
\ead{youssef.diouane@polymtl.ca}

\author[label8]{Jasper Bussemaker}
\ead{jasper.bussemaker@dlr.de}

\author[label3]{Thierry Lefebvre}
\ead{thierry.lefebvre@onera.fr}

\author[label2]{John T. Hwang}
\ead{jhwang@eng.ucsd.edu}


\author[label4,label7]{Joseph Morlier}
\ead{joseph.morlier@isae-supaero.fr}

\author[label1]{Joaquim R. R. A. Martins}
\ead{jrram@umich.edu}

\address[label3]{ONERA/DTIS, Université de Toulouse, Toulouse, France}
\address[label7]{ISAE-SUPAERO, Université de Toulouse, Toulouse, France}
\address[label6]{Polytechnique Montr\'eal, Montreal, QC, Canada}
\address[label8]{German Aerospace Center (DLR), Institute of System Architectures in Aeronautics, Hamburg, Germany}
\address[label2]{University of California San Diego, Department of Mechanical and Aerospace Engineering, La Jolla, CA, USA}
\address[label4]{ICA, Universit\'e de Toulouse, ISAE--SUPAERO, INSA, CNRS, MINES ALBI, UPS, Toulouse, France}
\address[label1]{University of Michigan, Department of Aerospace Engineering, Ann Arbor, MI, USA}

\begin{abstract}
The Surrogate Modeling Toolbox (SMT) is an open-source Python package that offers a collection of surrogate modeling methods, sampling techniques, and a set of sample problems.  
This paper presents \texttt{SMT 2.0}, a major new release of SMT that introduces significant upgrades and new features to the toolbox.
This release adds the capability to handle mixed-variable surrogate models and hierarchical variables.
These types of variables are becoming increasingly important in several surrogate modeling applications. 
\texttt{SMT 2.0} also improves SMT by extending sampling methods, adding new surrogate models, and computing variance and kernel derivatives for Kriging.
This release also includes new functions to handle noisy and use multi-fidelity data. 
To the best of our knowledge, \texttt{SMT 2.0} is the first open-source surrogate library to propose surrogate models for hierarchical and mixed inputs. 
This open-source software is distributed under the New BSD license~\footnote{\url{https://github.com/SMTorg/SMT}}.
%

\end{abstract}


\end{frontmatter}

\section{Motivation and significance}
\label{sec:intro}
With the increasing complexity and accuracy of numerical models, it has become more challenging to run complex simulations and computer codes~\cite{Mader_ad,Kennedy_Bayes}. 
As a consequence, surrogate models have been recognized as a key tool for engineering tasks such as design space exploration, uncertainty quantification, and optimization~\cite{Hwang2018b}. 
In practice, surrogate models are used to reduce the computational effort of these tasks by replacing expensive numerical simulations with closed-form approximations~\cite[Ch.~10]{Martins2021}.
To build such a model, we start by evaluating the original expensive simulation at a set of points through a \textcolor{black}{Design of Experiments (DoE)}.
Then, the corresponding evaluations are used to build the surrogate model according to the chosen approximation, such as Kriging, quadratic interpolation, or least squares regression.

The Surrogate Modeling Toolbox (SMT) is an open-source framework that provides functions to efficiently build surrogate models~\cite{SMT2019}.
Kriging models (also known as Gaussian processes) that take advantage of derivative information are one of SMT's key features~\cite{bouhlel2019gradient}.
Numerical experiments have shown that SMT achieved lower prediction error and computational cost than Scikit-learn~\cite{scikit-learn} and UQLab~\cite{UQLab} for a fixed number of points~\cite{ToolSMT}.
SMT has been applied to rocket engine coaxial-injector optimization~\cite{DL1}, aircraft engine consumption modeling~\cite{DGP1}, numerical integration~\cite{eliavs2020periodic}, multi-fidelity sensitivity analysis~\cite{drouet2023multi}, high-order robust finite elements methods~\cite{karban2021fem,kudela2022recent}, planning for photovoltaic solar energy~\cite{chen2020surrogate}, wind turbines design optimization~\cite{jasa2022effectively}, porous material optimization for a high pressure turbine vane~\cite{wang2023transpiration}, chemical process design~\cite{savage2020adaptive} and many other applications.
 
In systems engineering, architecture-level choices significantly influence the final system performance, and therefore, it is desirable to consider such choices in the early design phases~\cite{chan2022trying}. 
Architectural choices are parameterized with discrete design variables; examples include the selection of technologies, materials, component connections, and number of instantiated elements.
When design problems include both discrete variables and continuous variables, they are said to have \emph{mixed variables}.

When architectural choices lead to different sets of design variables, we have \emph{hierarchical} variables~\cite{Hutter,Architecture}.
For example, consider different aircraft propulsion architectures~\cite{fouda2022automated}.
A conventional gas turbine would not require a variable to represent a choice in the electrical power source, while hybrid or pure electric propulsion would require such a variable.
The relationship between the choices and the sets of variables can be represented by a hierarchy.

Handling hierarchical and mixed variables requires specialized  surrogate modeling techniques~\cite{Effectiveness}.
To address these needs, \texttt{SMT 2.0} is offering researchers and practitioners a collection of cutting-edge tools to build surrogate models with continuous, mixed and hierarchical variables.
The main objective of this paper is to detail the new enhancements that have been added in this release compared to the original \texttt{SMT 0.2} release~\cite{SMT2019}.

There are two new major capabilities in \texttt{SMT 2.0}: the ability to build surrogate models involving mixed variables and the support for hierarchical variables within Kriging models.
To handle mixed variables in Kriging models, existing libraries such as BoTorch~\cite{balandat2020botorch}, Dakota~\cite{Dakota}, DiceKriging~\cite{DiceKriging}, LVGP~\cite{zhang2020latent}, Parmoo~\cite{parmoo}, and Spearmint~\cite{GMHL} implement simple mixed models by using either continuous relaxation (CR), also known as \emph{one-hot encoding}~\cite{GMHL}, or a Gower distance (GD) based correlation kernel~\cite{Gower}. 
KerGP~\cite{Roustant} (developed in R) implements more general kernels but there is no Python open-source toolbox that implements more general kernels to deal with mixed variables, such as the homoscedastic hypersphere (HH)~\cite{Zhou} and exponential homoscedastic hypersphere (EHH)~\cite{Mixed_Paul} kernels.
Such kernels require the tuning of a large number of hyperparameters but lead to more accurate Kriging surrogates than simpler mixed kernels~\cite{Pelamatti,Mixed_Paul}. 
\texttt{SMT 2.0} implements all these kernels (CR, GD, HH, and EHH) through a unified framework and implementation. 
%
%
To handle hierarchical variables, no library in the literature can  build peculiar surrogate models except \texttt{SMT 2.0}, which implements two Kriging methods for these variables. 
Notwithstanding, most softwares are compatible with a naïve strategy called the imputation method~\cite{Effectiveness}   
but this method lacks depth and depends on arbitrary choices.
This is why~\citet{Hutter} proposed a first kernel, called \texttt{Arc-Kernel} which in turn was generalized by~\citet{Horn_hier} with a new kernel called the \texttt{Wedge-Kernel}~\cite{DACE_hier}.
None of these kernels are available in any open-source \textcolor{black}{modeling} software. Furthermore, thanks to the framework introduced in~\citet{audet2022general}, our proposed kernels are sufficiently general so that all existing hierarchical kernels are included within it. Section 4 describes the two kernels implemented in \texttt{SMT 2.0} that are referred as \texttt{SMT Arc-Kernel} and \texttt{SMT Alg-Kernel}.
In particular, \texttt{Alg-Kernel} is a novel hierarchical kernel introduced in this paper. 
%
%
Table~\ref{tab:comparison} outlines the main features of the state-of-the-art modeling software that can handle hierarchical and mixed variables.
\begin{table}[H]
\caption{Comparison of software packages for hierarchical and mixed Kriging models. \checkmark = implemented. * = user-defined. 
}
\resizebox{\columnwidth}{!}{%
\begin{tabular}{l c c c c c c c c }
\hline
\textbf{Package} & \texttt{BOTorch} & \texttt{Dakota} & \texttt{DiceKriging} & \texttt{KerGP} & \texttt{LVGP} & \texttt{Parmoo}  & \texttt{Spearmint} & \texttt{SMT 2.0} \\
\hline
\textbf{Reference} & \cite{balandat2020botorch} & \cite{Dakota} & \cite{DiceKriging} & \cite{Roustant} &\cite{zhang2020latent}  &\cite{parmoo} &\cite{GMHL} & This paper    \\
\textbf{License}   & MIT & EPL & GPL & GPL & GPL & BSD &  GNU & BSD  \\
\textbf{Language}  & Python & C & R & R & R & Python & Python &  Python \\

\textbf{Mixed var.} & \checkmark & \checkmark & \checkmark & \checkmark & \checkmark & \checkmark & \checkmark & \checkmark \\
\textit{GD kernel} & \checkmark & \checkmark & \checkmark & *  & &  &  & \checkmark \\
\textit{CR kernel} &  &  &  & & \checkmark  &  \checkmark  &  \checkmark & \checkmark\\
\textit{HH kernel} &  &  & & \checkmark  &  &  &  & \checkmark  \\
\textit{EHH kernel} &  &  &  & *  &  & &   & \checkmark \\
\textbf{Hierarchical var.} &  &  &  &  &  & &   & \checkmark \\
\hline
\end{tabular}
}
\label{tab:comparison}
\end{table}


\texttt{SMT 2.0} introduces other enhancements, such as additional sampling procedures, new surrogate models, new Kriging kernels (and their derivatives), Kriging variance derivatives, and an adaptive criterion for high-dimensional problems. 
\texttt{SMT 2.0} adds applications of Bayesian optimization (BO) with hierarchical and mixed variables or noisy co-Kriging that have been successfully applied to aircraft design~\cite{SciTech_cat}, data fusion~\cite{condearenzana}, and structural design~\cite{RaulAIAA}.
The \texttt{SMT 2.0} interface is more user-friendly and offers an improved and more detailed documentation for users and developers~\footnote{\url{http://smt.readthedocs.io/en/latest}}. 
\texttt{SMT 2.0} is hosted publicly~\footnote{\url{https://github.com/SMTorg/smt}} and can be directly imported within Python scripts.
It is released under the New BSD License and runs on Linux, MacOS, and Windows operating systems.
Regression tests are run automatically for each operating system whenever a change is committed to the repository.
In short, \texttt{SMT 2.0} builds on the strengths of the original SMT package while adding new features. On one hand, the emphasis on derivatives (including prediction, training and output derivatives) is maintained and improved in \texttt{SMT 2.0}. On the other hand, this new release includes support for hierarchical and mixed variables Kriging based models.
For the sake of reproducibility, an open-source notebook is available that gathers all the methods and results presented on this paper~\footnote{\url{https://github.com/SMTorg/smt/tree/master/tutorial/NotebookRunTestCases_Paper_SMT_v2.ipynb}}.

The remainder of the paper is organized as follows.
First, we introduce the organization and the main implemented features of the release in Section~\ref{sec:description}. 
Then, we describe the mixed-variable Kriging model with an example in Section~\ref{sec:mixed}. 
Similarly, we describe and provide an example for a hierarchical-variable Kriging model in Section~\ref{sec:hier}.
The Bayesian optimization models and applications are described in  Section~\ref{sec:BO}.
Finally, we describe  the other relevant contributions in Section~\ref{sec:other} and conclude in Section~\ref{sec:concl}.

\section{\texttt{SMT 2.0}: an improved surrogate modeling toolbox}
\label{sec:description}

From a software point of view, \texttt{SMT 2.0} maintains and improves the modularity and generality of the original SMT version~\cite{SMT2019}. 
In this section, we describe the software as follows.
Section~\ref{sec:legacy} describes the legacy of \texttt{SMT 0.2}.
Then, Section~\ref{sec:organization} describes the organization of the repository.
Finally, Section~\ref{sec:newSMT2} shows the new capabilities implemented in the \texttt{SMT 2.0} update.

\subsection{Background on SMT former version: \texttt{SMT 0.2} }
\label{sec:legacy}  

SMT~\cite{SMT2019} is an open-source collaborative work originally developed by ONERA, NASA Glenn, ISAE-SUPAERO/ICA and the University of Michigan.
Now, both Polytechnique Montréal and the University of California San Diego are also contributors. 
\texttt{SMT 2.0} updates and extends the original SMT repository capabilities among which the original publication~\cite{SMT2019} focuses on different types of derivatives for surrogate models detailed hereafter.

\paragraph{A Python surrogate modeling framework with derivatives}
One of the original main motivations for SMT was derivative support. In fact, none of the existing packages for surrogate modeling such as Scikit-learn in Python~\cite{scikit-learn}, SUMO in Matlab~\cite{Gorissen2010} or GPML in Matlab and Octave~\cite{williams2006gaussian} focuses on derivatives. 
Three types of derivatives are distinguished: prediction derivatives, training derivatives, and output derivatives.
SMT also includes new models with derivatives such as Kriging with Partial Least Squares (KPLS)~\cite{Bouhlel18} and Regularized Minimal-energy Tensor-product Spline (RMTS)~\cite{Hwang2018b}. 
These developed derivatives were even used in a novel algorithm called Gradient-Enhanced Kriging with Partial Least Squares (GEKPLS)~\cite{bouhlel2019gradient} to use with adjoint methods, for example~\cite{bouhlel2020}. 

\paragraph{Software architecture, documentation, and automatic testing}
SMT is organized along three main sub-modules that implement a set of sampling techniques (\textbf{sampling\_methods}), benchmarking functions (\textbf{problems}), and surrogate modeling techniques (\textbf{surrogate\_models}). 
The toolbox documentation~\footnote{\url{https://smt.readthedocs.org}} is created using reStructuredText and Sphinx, a documentation generation package for Python, with custom extensions.
Code snippets in the documentation pages are taken directly from actual tests in the source code and are automatically updated. 
The output from these code snippets and tables of options are generated dynamically by custom Sphinx extensions. 
This leads to high-quality documentation with minimal effort.
Along with user documentation, developer documentation is also provided to explain how to contribute to SMT.
This includes a list of API methods for the \textbf{SurrogateModel}, \textbf{SamplingMethod}, and \textbf{Problem} classes, that must be implemented to create a new surrogate modeling method, sampling technique, or benchmarking problem.
When a developer submits a pull request, it is merged only after passing the automated tests and receiving approval from at least one reviewer.
The repository on GitHub~\footnote{\url{https://github.com/SMTorg/smt}} is linked to continuous integration tests (GitHub Actions) for Windows, Linux and MacOS, to a coverage test on coveralls.io and to a dependency version check for Python with DependaBot.
Various parts of the source code have been accelerated using Numba~\cite{numba}, an LLVM-based just-in-time (JIT) compiler for numpy-heavy Python code. Numba is applied to conventional Python code using function decorators, thereby minimizing its impact on the development process and not requiring an additional build step. 
For a mixed Kriging surrogate with 150 training points, a speedup of up to 80\% is observed, see Table~\ref{tab:numba_benchmark}.
The JIT compilation step only needs to be done once when installing or upgrading SMT and adds an overhead of approximately 24 seconds on a typical workstation.
\textcolor{black}{
In this paper, all results are obtained using an Intel® Xeon® CPU E5-2650 v4 @ 2.20 GHz core and 128 GB of memory with a Broadwell-generation processor front-end and a compute node of a peak power of 844 GFlops.
}

\begin{table}[H]
\caption{Impact of using Numba on training time of the hierarchical Goldstein problem. Speedup is calculated excluding the JIT compilation table, as this step is only needed once after SMT installation.}
\begin{center}
\resizebox{\linewidth}{!}{%
\begin{tabular}{lcccc}
\hline
\textbf{Training set} & Without Numba & Numba & Speedup & JIT overhead \\
\hline
\textbf{15 points} & 1.3 s & 1.1 s & 15\% & 24 s \\
\textbf{150 points} & 38 s & 7.4 s & 80\% & 23 s \\
\hline
\end{tabular}
}
\end{center}
\label{tab:numba_benchmark}
\end{table}

\subsection{Organization of \texttt{SMT 2.0}} 
\label{sec:organization}
\begin{figure}[h!]
\centering
\vspace{3pt}
\begin{forest}
      for tree={
        font= \small \ttfamily,
        grow'=0,
        child anchor=west,
        parent anchor=south,
        anchor=west,
        calign=first,
        inner xsep=10pt,
        inner ysep = -3pt,
        edge path={
          \noexpand\path [draw, \forestoption{edge}]
          (!u.south west) +(7.5pt,0) |- (.child anchor)  \forestoption{edge label};
        },
        file/.style={edge path={\noexpand\path [draw, \forestoption{edge}]
          (!u.south west) +(7.5pt,0) |- (.child anchor) \forestoption{edge label};},
          inner xsep=2pt, inner ysep=-3pt ,font=\footnotesize \ttfamily
                     },
        before typesetting nodes={
          if n=1
            {insert before={[,phantom]}}
            {}
        },
        fit=band,
        before computing xy={l=15pt},
      }  
    [ \   \normalsize \texttt{SMT 2.0}
      [ \textcolor{blue}{Sampling methods}
        [Random, file
        ]      
        [Full Factorial, file
        ]
        [ Latin Hypercube Sampling \textcolor{Orchid}{(Nested LHS and Extended LHS)}, file
        ]
      ]
      [ \textcolor{blue}{Problems}
        [Aircraft wing weight, file
        ]      
        [Robot arm position, file
        ]  
        [Water flow through a borehole, file
        ]   
        [Low frequency torsion vibration, file
        ]    
        [Welded beam shear stress, file
        ]  
        [ \textcolor{Orchid}{Mixed integer cantilever beam}, file
        ]
        [ \textcolor{Orchid}{Hierarchical neural network}, file
        ]      
      ]
      [ \textcolor{blue}{Surrogate modeling methods}
        [ RBF: Radial Basis Functions, file
        ]
        [IDW: Inverse-Distance Weighting, file
        ]
        [RMTS: Regularized Minimal-energy Tensor-product Splines, file
        ]
         [LS: Least-Squares approximation, file
        ]      
         [QP: Quadratic Polynomial approximation,  file
        ]
        [Kriging based models,  file
            [Continuous kernels,  file]
            [\textcolor{Orchid}{Hierarchical kernels},  file]
            [\textcolor{Orchid}{Categorical kernels},  file]
        ]
        [\textcolor{Orchid}{GENN: Gradient-Enhanced Neural Network},  file ]
         [ \textcolor{Orchid}{MGP: Marginal Gaussian Process},  file ]
      ]
      [ \textcolor{blue}{Applications}
          [   Mixture of experts (MOE) , file]
          [  Variable-fidelity modeling (VFM) , file]
          [  Multi-fidelity Kriging (MFK) , file]
         [    \textcolor{Orchid}{ Multi-fidelity KPLS (MFKPLS)} , file]
         [    \textcolor{Orchid}{ Multi-fidelity KPLSK (MFKPLSK)} , file]
         [   Efficient Global Optimization (EGO) , file]
         [  \textcolor{Orchid}{ Mixed-Integer and hierarchical usage surrogates}, file]
      ]
        [ \textcolor{blue}{Interactive notebooks}
          [  SMT tutorial for surrogate modeling, file]
          [  \textcolor{Orchid}{  Noisy Gaussian process (Kriging) }, file]
          [ \textcolor{Orchid}{Multi-fidelity Gaussian process (with or without noise)} , file]
         [  \textcolor{Orchid}{ Gaussian process trajectory Sampling} , file]
         [ \textcolor{Orchid}{ Bayesian optimization to solve unconstrained problems} , file]
         [ \textcolor{Orchid}{ Mixed \& hierarchical Kriging and optimization}, file]
         [ \textcolor{Orchid}{SMT 2.0 Advancements and hierarchical variables}, file]
      ]
    ]
 \end{forest}

\caption{\label{fig:smt_codes} Functionalities of \texttt{SMT 2.0}. 
The new major features implemented in \texttt{SMT 2.0} compared to \texttt{SMT 0.2} are highlighted with the lavender color.}
\end{figure}
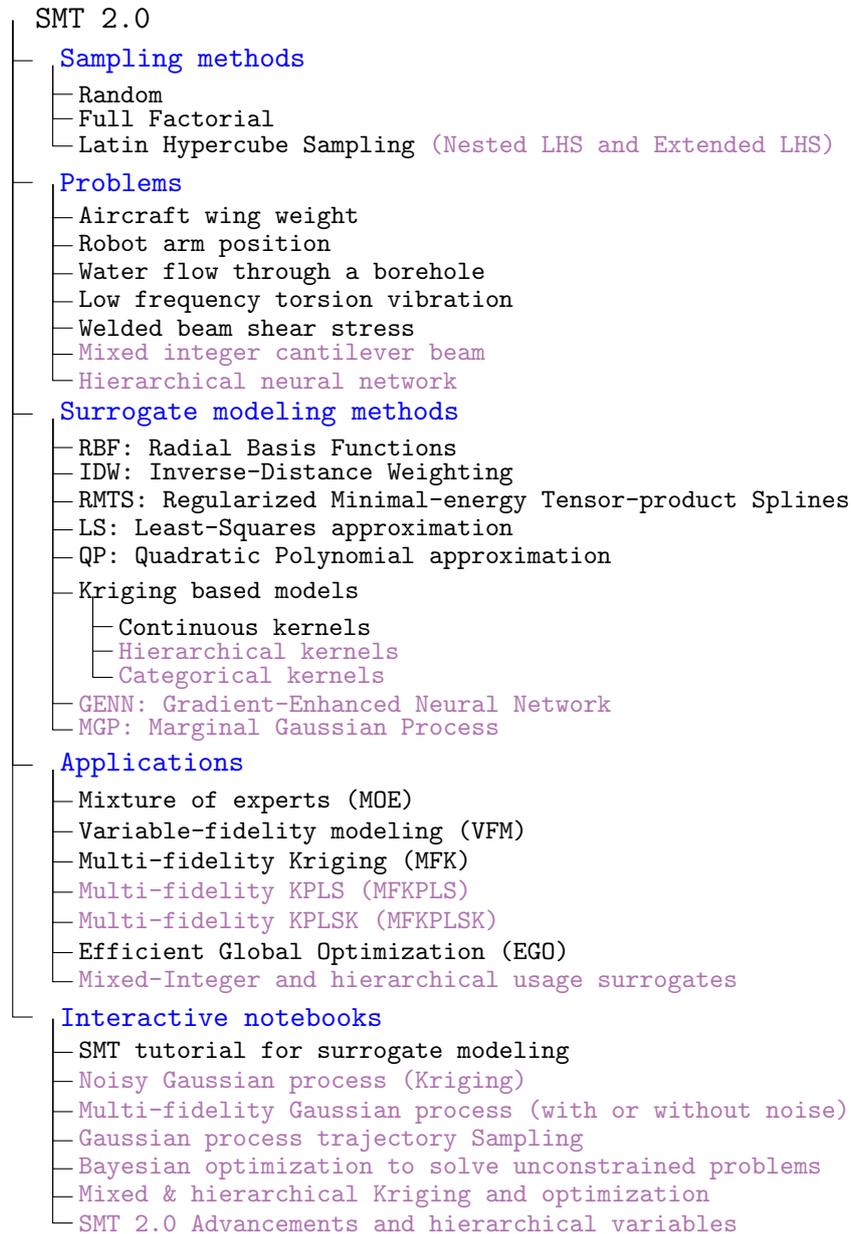

The main features of the open-source repository \texttt{SMT 2.0} are described in~\figref{fig:smt_codes}. 
More precisely, \texttt{Sampling Methods}, \texttt{Problems} and \texttt{Surrogate models} are kept from \texttt{SMT 0.2} and two new sections \texttt{Models applications} and \texttt{Interactive notebooks} have been added to the architecture of the code. 
These sections are highlighted in blue and detailed on~\figref{fig:smt_codes}. 
The new major features implemented in \texttt{SMT 2.0} are highlighted in lavender whereas the legacy features that were already in present in the original publication for \texttt{SMT 0.2}~\cite{SMT2019} are in black.

\subsection{New features within \texttt{SMT 2.0} } 
\label{sec:newSMT2}
The main objective of this new release is to enable Kriging surrogate models for use with both hierarchical and mixed variables.
Moreover, for each of these five sub-modules described in Section~\ref{sec:organization}, several improvements have been made between the original version and the \texttt{SMT 2.0} release.

\paragraph{Hierarchical and mixed design space}
A new design space definition class \texttt{DesignSpace} has been added that implements hierarchical and mixed functionalities. 
Design variables can either be continuous ({\footnotesize \texttt{FloatVariable}}), ordered ({\footnotesize \texttt{OrdinalVariable}}) or categorical ({\footnotesize \texttt{CategoricalVariable}}). 
The integer type ({\footnotesize \texttt{IntegerVariable}}) represents a special case of the ordered variable, specified by bounds (inclusive) rather than a list of possible values.
The hierarchical structure of the design space can be defined using {\footnotesize \texttt{declare\_decreed\_var}}: this function declares that a variable is a decreed variable that is activated when the associated meta variable takes one of a set of specified values, see Section~\ref{sec:hier} for background.
The \texttt{DesignSpace} class also implements mechanisms for sampling valid design vectors (i.e. design vectors that adhere to the hierarchical structure of the design space) using any of the below-mentioned samplers, for correcting and imputing design vectors, and for requesting which design variables are acting in a given design vector. 
Correction ensures that variables have valid values (e.g. integers for discrete variables)~\cite{Effectiveness}, and imputation replaces non-acting variables by some default value ($0$ for discrete variables, mid-way between the bounds for continuous variables in \texttt{SMT 2.0})~\cite{Zaefferer}.

\paragraph{Sampling} 
SMT implements three methods for sampling.
The first one is a naïve approach, called \texttt{Random} that draws uniformly points along every dimension.
The second sampling method is called \texttt{Full Factorial} and draws a point for every cross combination of variables, to have an "exhaustive" design of experiments. 
The last one is the \texttt{Latin Hypercube Sampling} (LHS)~\cite{LHS} that draws a point in every Latin square parameterized by a certain criterion. 
For LHS, a new criterion to manage the randomness has been implemented and the sampling method was adapted for multi-fidelity and mixed or hierarchical variables.
More details about the new sampling techniques are given in Section~\ref{sec:sampling}.

\paragraph{Problems} 
SMT implements two new engineering problems: a mixed variant of a cantilever beam described in Section~\ref{sec:mixed} and a hierarchical neural network described in Section~\ref{sec:hier}.

\paragraph{Surrogate models} 
In order to keep up with state-of-art, several releases done from the original version developed new options for the already existing surrogates.
In particular, compared to the original publication~\cite{SMT2019}, \texttt{SMT 2.0} adds gradient-enhanced neural networks~\cite{bouhlel2020} and marginal Gaussian process~\cite{MGP} models to the list of available surrogates. 
More details about the new models are given in  Section~\ref{sec:surrogates}.

\paragraph{Applications} 
Several applications have been added to the toolbox to demonstrate the surrogate models capabilities. 
The most relevant application is efficient global optimization (EGO), a Bayesian optimization algorithm~\cite{Jones2001JOGO,lafage2022egobox}. 
EGO optimizes expensive-to-evaluate black-box problems with a chosen surrogate model and a chosen optimization criterion~\cite{Jones98}.
The usage of EGO with hierarchical and mixed variables is described in Section~\ref{sec:BO}.

\paragraph{Interactive notebooks} 
These tutorials introduce and explain how to use the toolbox for different surrogate models and applications~\footnote{\url{ https://github.com/SMTorg/smt/tree/master/tutorial}}.
Every tutorial is available both as a \texttt{.ipynb} file and directly on Google colab~\footnote{\url{https://colab.research.google.com/github/SMTorg/smt/ }}. 
In particular, a hierarchical and mixed variables dedicated notebook is available to reproduce the results presented on this paper~\footnote{\url{https://github.com/SMTorg/smt/tree/master/tutorial/NotebookRunTestCases_Paper_SMT_v2.ipynb}}.

In the following, Section~\ref{sec:mixed} details the Kriging based surrogate models for mixed variables, and Section~\ref{sec:hier} presents our new Kriging surrogate for hierarchical variables.
Section~\ref{sec:BO} details the EGO application and the other new relevant features aforementioned are described succinctly in Section~\ref{sec:other}.

\section{Surrogate models with mixed variables in \texttt{SMT 2.0}}
\label{sec:mixed}

As mentioned in Section~\ref{sec:intro}, design variables can be either of continuous or discrete type, and a problem with both types is a mixed-variable problem.
Discrete variables can be ordinal or categorical.  
A discrete variable is \emph{ordinal} if there is an order relation within the set of possible values. 
An example of an ordinal design variable is the number of engines in an aircraft.
A possible set of values in this case could be ${2, 4, 8}$.
A discrete variable is \emph{categorical} if no order relation is known between the possible choices the variable can take. 
One example of a categorical variable is the color of a surface.
A possible example of a set of choices could be ${\text{blue}, \text{red}, \text{green}}$.
The possible choices are called the \emph{levels} of the variable.


Several methods have been proposed to address the recent increase interest in mixed Kriging based models~\cite{Pelamatti, Zhou, Deng, Roustant,GMHL,Gower,cuesta2021comparison,SciTech_cat}. 
The main difference from a continuous Kriging model is in the estimation of the categorical correlation matrix, which is critical to determine the mean and variance predictions.
As mentioned in Section~\ref{sec:intro}, approaches such as CR~\cite{GMHL,SciTech_cat}, continuous latent variables~\cite{cuesta2021comparison}, and GD~\cite{Gower} use a kernel-based method to estimate the correlation matrix.
Other methods estimate the correlation matrix by modeling the correlation entries directly~\cite{Pelamatti, Deng, Roustant}, such as HH~\cite{Zhou} and EHH~\cite{Mixed_Paul}. 
The HH correlation kernel is of particular interest because it generalizes simpler kernels such as CR and GD~\cite{Mixed_Paul}.
In \texttt{SMT 2.0}, the correlation kernel is an option that can be set to either CR (\texttt{CONT\_RELAX\_KERNEL}), GD ( {\texttt{GOWER\_KERNEL}), HH (\texttt{HOMO\_HSPHERE\_KERNEL}) or EHH (\texttt{EXP\_HOMO\_HSPHERE\_KERNEL}).

\subsection{Mixed Gaussian processes}

The continuous and ordinal variables are both treated similarly in \texttt{SMT 2.0} with a continuous kernel, where the ordinal values are converted to continuous through relaxation. 
For categorical variables, four models (GD, CR, EHH and HH) can be used in \texttt{SMT 2.0} if specified by the API. 
This is why we developed a unified mathematical formulation that allows a unique implementation for any model.   

Denote $l$ the number of categorical variables.
For a given  $i \in \{1, \ldots, l\}$, the $ i^{\text{th}}$ categorical variable is denoted $c_i$ and its number of levels is denoted $L_i$. 
The hyperparameter matrix peculiar to this variable $c_i$ is 
$$\Theta_i= \begin{bmatrix}
[\Theta_i]_{1,1} & \textcolor{white}{9} & \hspace{2em} { \textbf{\textit{ Sym.}}}  \textcolor{white}{9} & \\
[\Theta_i]_{1,2}  & [\Theta_i]_{2,2} & \textcolor{white}{9} \\
\vdots &\ddots & \ddots & \textcolor{white}{9}  \\
[\Theta_i]_{1,L_i} &  \ldots & [\Theta_i]_{L_i-1,L_i} &[\Theta_i]_{L_i,L_i} \\ 
\end{bmatrix}, $$
and the categorical parameters are defined as $\theta^{cat} = \{ \Theta_1 , \ldots, \Theta_l \}$.
For two given inputs in the DoE, for example, the $r^{\text{th}}$ and  $s^{\text{th}}$ points, let $c^r_{i} $ and $c^s_{i} $ be the associated categorical variables taking respectively the $\ell^i_r$ and the $\ell^i_s$ level on the categorical variable $c_i$. The categorical correlation kernel is defined by
\begin{equation}
\begin{split}
&k^{cat}(c^r,c^s,\theta^{cat}) = {\displaystyle \prod_{i=1}^{l}   \kappa ( 2 [ \Phi(\Theta_i) ]_{{ \ell_i^s},{\ell_i^r}} ) \  \kappa ( [ \Phi(\Theta_i) ]_{{ \ell_i^r},{\ell_i^r}} ) \  \kappa ( [ \Phi(\Theta_i) ]_{{ \ell_i^s},{\ell_i^s}} ) }
\end{split}
\end{equation}
where $\kappa$ is either a positive definite kernel or identity and $\Phi(.)$ is a symmetric positive definite (SPD) function such that the matrix $\Phi(\Theta_i) $ is SPD if $\Theta_i$ is SPD.
For an exponential kernel,~\tabref{tab:kernels} gives the parameterizations of $\Phi$ and $\kappa$ that correspond to GD, CR, HH, and EHH kernels. 
The complexity of these different kernels depends on the number of hyperparameters that characterizes them. 
As defined by \citet{Mixed_Paul}, for every categorical variable $i \in \{1, \ldots, l\}$, the matrix $C(\Theta_i)\in \mathbb{R}^{L_i \times L_i}$ is lower triangular and built using a hypersphere decomposition~\cite{HS,HS_Jacobi} from the symmetric matrix $\Theta_i \in \mathbb{R}^{L_i \times L_i}$ of hyperparameters. 
The variable $\epsilon$ is a small positive constant and the variable $\theta_{i}$ denotes the only positive hyperparameter that is used for the Gower distance kernel.

\begin{table}[H]
\caption{Categorical kernels implemented in \texttt{SMT 2.0}.}
\vspace{-0.5cm}
\begin{center}
\resizebox{\columnwidth}{!}{%
\begin{tabular}{cclc}
\hline
\textbf{Name} & $\kappa(\phi)$   &  \hspace{3cm} ${\centering \Phi(\Theta_i)}$   &  \# of hyperparam. \\
\hline
\texttt{SMT GD}   &  $\exp(-\phi) $ & ${ \displaystyle [\Phi(\Theta_i)]_{j,j} :=  \frac{1}{2}  \theta_{i} \quad  ~;~ [\Phi(\Theta_i)]_{j \neq j'} := 0 }$ & 1   \\
\texttt{SMT CR}  & $\exp(-\phi) $ &  $  { \displaystyle [\Phi(\Theta_i)]_{j,j} := [\Theta_i]_{j,j} ~;~ [\Phi(\Theta_i)]_{j \neq j'} := 0 } $  & $L_i$  \\
\texttt{SMT EHH}  & $\exp(-\phi)$ & 
$  { \displaystyle [\Phi(\Theta_i)]_{j,j} := 0 \quad \quad ~;~ [\Phi(\Theta_i)]_{j \neq j'} := \frac{\log \epsilon }{2} ([C(\Theta_i) C(\Theta_i) ^\top]_{j,j'} -1)  }$  & $\frac{1}{2}  (L_i)  (L_i-1) $\\
\texttt{SMT HH}  &  $\phi$ &    $  { \displaystyle [\Phi(\Theta_i)]_{j,j} := 1 \quad \quad ~;~ [\Phi(\Theta_i)]_{j \neq j'} :=  \frac{1}{2}  \textcolor{black}{[C(\Theta_i) C(\Theta_i)^\top]_{j,j'} }}$ & $\frac{1}{2}  (L_i)  (L_i-1) $  \\
\hline
\end{tabular}
}
\end{center}
\label{tab:kernels}
\end{table}

Another Kriging based model that can use mixed variables is Kriging with partial least squares (KPLS)~\cite{bouhlel_KPLSK}.
KPLS adapts Kriging to high dimensional problems by using a reduced number of hyperparameters thanks to a projection into a smaller space. 
Also, for a general surrogate, not necessarily Kriging, \texttt{SMT 2.0} uses continuous relaxation to allow whatever model to handle mixed variables. 
For example, we can use mixed variables with least squares (LS) or quadratic polynomial (QP) models.
We now illustrate the abilities of the toolbox in terms of mixed modeling over an engineering test case.

\subsection{An engineering design test-case}
\label{sec:beam}
A classic engineering problem commonly used for model validation is the beam bending problem~\cite{Roustant, Cheng2015TrustRB}.
This problem is illustrated on~\figref{fig:beam} and consists of a cantilever beam in its linear range loaded at its free end with a force $F$. 
As in~\citet{Cheng2015TrustRB}, the Young modulus is  $E=200$GPa and the chosen load is $F=50$kN.
Also, as in~\citet{Roustant}, 12 possible cross-sections can be used. These 12 sections consist of 4 possible shapes that can be either hollow, thick or full as illustrated in~\figref{fig:beam_shape}.

\begin{figure}[ht]
\centering

\vspace{-5pt}
\captionsetup{justification=raggedright,singlelinecheck=false}
\subfloat[Bending problem.]{
\begin{tikzpicture}
    \hspace{-3pt}
    \point{origin}{-0.75}{-0.25};
    \point{begin}{0}{0};
    \point{end}{5}{0};
    \point{end_bot}{4.99}{-0.9};
    \point{end_up}{5}{0.5};
    \beam{2}{begin}{end};
    \support{3}{begin}[-90];
    \load{1}{end}[90]   ;
    \notation{1}{end_up}{$F=50kN$};

     \draw[<->] (end) -- (end_bot) node[midway, right] {$\delta$} ;
     \draw[<->] (0,0.5) -- (5,0.5) node[midway, above] {L};
     
    \draw
      [-, ultra thick] (begin) .. controls (1.5, +.01) and (2.5, -.15) .. (4.93, -0.9)
      [-, ultra thick] (begin) .. controls (1.5, +.01) and (2.5, -.2) .. (4.85, -1.5)
      [-, ultra thick] (begin) .. controls (1.5, +.01) and (2.5, -.4)   .. (4.78, -1.9);
  \end{tikzpicture}
\label{fig:beam}   
}
\subfloat[Possible cross-section shapes.]{
\centering
\begin{tikzpicture}
\hspace{-50pt}

\tstar{0.25}{0.5}{6}{0}{thick,fill=yellow}
\tstar{0.14}{0.28}{6}{0}{thick,fill=white}

\tstar{0.25}{0.5}{6}{0}{thick,fill=yellow,xshift=-1.2cm}
\tstar{0.08}{0.16}{6}{0}{thick,fill=white,xshift=-1.2cm}

\tstar{0.25}{0.5}{6}{0}{thick,fill=yellow,xshift=-2.4cm}

\def\pos{1.2}
\fill[black] (\pos-0.24,-0.5) -- (\pos-0.24,0.5) -- (\pos+0.24,0.5)  -- (\pos+0.24,-0.5)   -- cycle ;
\fill[black] (\pos-0.5,-0.5) -- (\pos-0.5,-0.18) -- (\pos+0.5,-0.18)  -- (\pos+0.5,-0.5)   -- cycle ; 
\fill[black] (\pos-0.5,0.18) -- (\pos-0.5,0.5) -- (\pos+0.5,0.5)  -- (\pos+0.5,0.18)   -- cycle ;  

\def\pos{2.4}
\fill[black] (\pos-0.19,-0.5) -- (\pos-0.19,0.5) -- (\pos+0.19,0.5)  -- (\pos+0.19,-0.5)   -- cycle ;
\fill[black] (\pos-0.5,-0.5) -- (\pos-0.5,-0.25) -- (\pos+0.5,-0.25)  -- (\pos+0.5,-0.5)   -- cycle ; 
\fill[black] (\pos-0.5,0.25) -- (\pos-0.5,0.5) -- (\pos+0.5,0.5)  -- (\pos+0.5,0.25)   -- cycle ;  

\def\pos{3.6}
\fill[black] (\pos-0.14,-0.5) -- (\pos-0.14,0.5) -- (\pos+0.14,0.5)  -- (\pos+0.14,-0.5)   -- cycle ;
\fill[black] (\pos-0.5,-0.5) -- (\pos-0.5,-0.32) -- (\pos+0.5,-0.32)  -- (\pos+0.5,-0.5)   -- cycle ; 
\fill[black] (\pos-0.5,0.32) -- (\pos-0.5,0.5) -- (\pos+0.5,0.5)  -- (\pos+0.5,0.32)   -- cycle ;  

\vspace{1.2cm}

\fill[green,even odd rule] (0,-1.2) circle (0.5) (0,-1.2) circle (0.33);
\draw (0,-1.2) circle (0.5) ;
\draw (0,-1.2) circle (0.33) 
; 
\fill[green,even odd rule] (-1.2,-1.2) circle (0.5)(-1.2,-1.2) circle (0.17);
\draw (-1.2,-1.2) circle (0.5) ;
\draw (-1.2,-1.2) circle (0.17) 
; 
\fill[green,even odd rule] (-2.4,-1.2) circle (0.5) ;
\draw (-2.4,-1.2) circle (0.5) ;

\def\pos{1.2}
\fill[blue,even odd rule]  (\pos-0.5,-1.7) -- (\pos-0.5,-0.7) -- (\pos+0.5,-0.7) -- (\pos+0.5,-1.7) -- cycle ;

\def\pos{2.4}
\fill[blue,even odd rule]  (\pos-0.5,-1.7) -- (\pos-0.5,-0.7) -- (\pos+0.5,-0.7) -- (\pos+0.5,-1.7) -- cycle   (\pos-0.25,-1.45) -- (\pos-0.25,-0.95) -- (\pos+0.25,-0.95) -- (\pos+0.25,-1.45) -- cycle ;

\def\pos{3.6}
\fill[blue,even odd rule]  (\pos-0.5,-1.7) -- (\pos-0.5,-0.7) -- (\pos+0.5,-0.7) -- (\pos+0.5,-1.7) -- cycle   (\pos-0.35,-1.55) -- (\pos-0.35,-0.85) -- (\pos+0.35,-0.85) -- (\pos+0.35,-1.55) -- cycle ;
  \end{tikzpicture}
  \label{fig:beam_shape}    
}
\captionsetup{justification=centering,singlelinecheck=false}
\caption{Cantilever beam problem~\cite[Figure 6]{Mixed_Paul}\textcolor{black}{.}}
\end{figure}
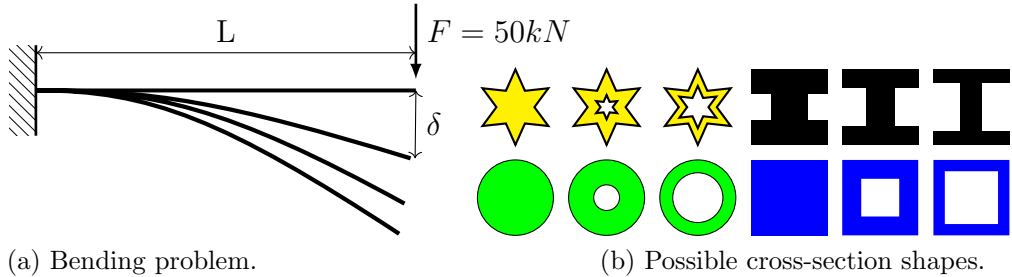

To compare the mixed Kriging models of \texttt{SMT 2.0}, we draw a 98 point LHS as training set and the validation set is a grid of $12\times30\times30=10800$ points. 
For the four implemented methods, displacement error (computed with a root-mean-square error criterion), likelihood, number of hyperparameters and computational time for every model are shown in~\tabref{tab:resCantilever}.
For the continuous variables, we use the square exponential kernel.
More details are found in~\cite{Mixed_Paul}.
As expected, the complex EHH and HH models lead to a lower displacement error and a higher likelihood value, but use more hyperparameters and increase the computational cost compared to GD and CR. 
On this test case, the kernel EHH is easier to optimize than HH but in general, they are similar in terms of performance. 
Also, by default \texttt{SMT 2.0} uses CR as it is known to be a good trade-off between complexity and performance~\cite{well-adapted_cont}.

\begin{table}[H]
\centering
\caption{Results of the cantilever beam models~\cite[Table 4]{Mixed_Paul}.}
\resizebox{\columnwidth}{!}{%
\begin{tabular}{ccccc}
\hline
\textbf{Categorical kernel} &  Displacement error (cm) & $ \ $ Likelihood  & \# of hyperparam.  
\\
\hline 
\texttt{SMT GD}   &1.3861 & 111.13&  3 
\\  
\texttt{SMT CR}   & 1.1671 & 155.32 & 14  
\\
\texttt{SMT EHH}   & {0.1613}   &
{236.25} & 68 
\\
\texttt{SMT HH}   & {0.2033} &
{235.66} & 68 
\\
\hline
\end{tabular}
}
\label{tab:resCantilever}
\end{table}

\section{Surrogate models with hierarchical variables in \texttt{SMT 2.0}}
\label{sec:hier}

To introduce the newly developed Kriging model for hierarchical variables implemented in \texttt{SMT 2.0}, we present the general mathematical framework for hierarchical and mixed variables established by \citet{audet2022general}.
In \texttt{SMT 2.0}, two variants of our new method are implemented, namely \texttt{SMT Alg-Kernel} and \texttt{SMT Arc-Kernel}.
In particular, the \texttt{SMT Alg-Kernel} is a novel correlation kernel introduced in this paper.

\subsection{The hierarchical variables framework}

A problem structure is classified as hierarchical when the sets of active variables depend on architectural choices.
This occurs frequently in industrial design problems.
In hierarchical problems, we can classify variables as neutral, meta (also known as dimensional) or decreed (also known as conditionally active) as detailed in~\citet{audet2022general}.
Neutral variables are the variables that are not affected by the hierarchy whereas the value assigned to meta variables determines which decreed variables are activated. For example, a meta variable could be the number of engines. If the number of engines changes, the number of decreed bypass ratios that every engine should specify also changes. However, the wing aspect ratio being neutral, it is not affected by this hierarchy. 

%

Problems involving hierarchical variables are generally dependant on discrete architectures and as such involve mixed variables. 
Hence, in addition to their role (neutral, meta or decreed), each variable also has a variable type amongst categorical, ordinal or continuous. 
For the sake of simplicity and because both continuous and ordinal variables are treated similarly~\cite{Mixed_Paul}, we chose to regroup them as \emph{quantitative variables}.
For instance, the neutral variables $x_{\neutral}$ may be partitioned into different variable types, such that $x_{\neutral} = (x_{\neutral}^{\cat},x_{\neutral}^{\quant})$ where $x_{\neutral}^{\cat}$ represents the categorical variables and $x_{\neutral}^{\quant}$ are the quantitative ones. 
The variable classification scheme in \texttt{SMT 2.0} is detailed in~\figref{fig:vars}.
%
\begin{figure}
\centering
\includegraphics[height=8.5cm,,width=14cm]{  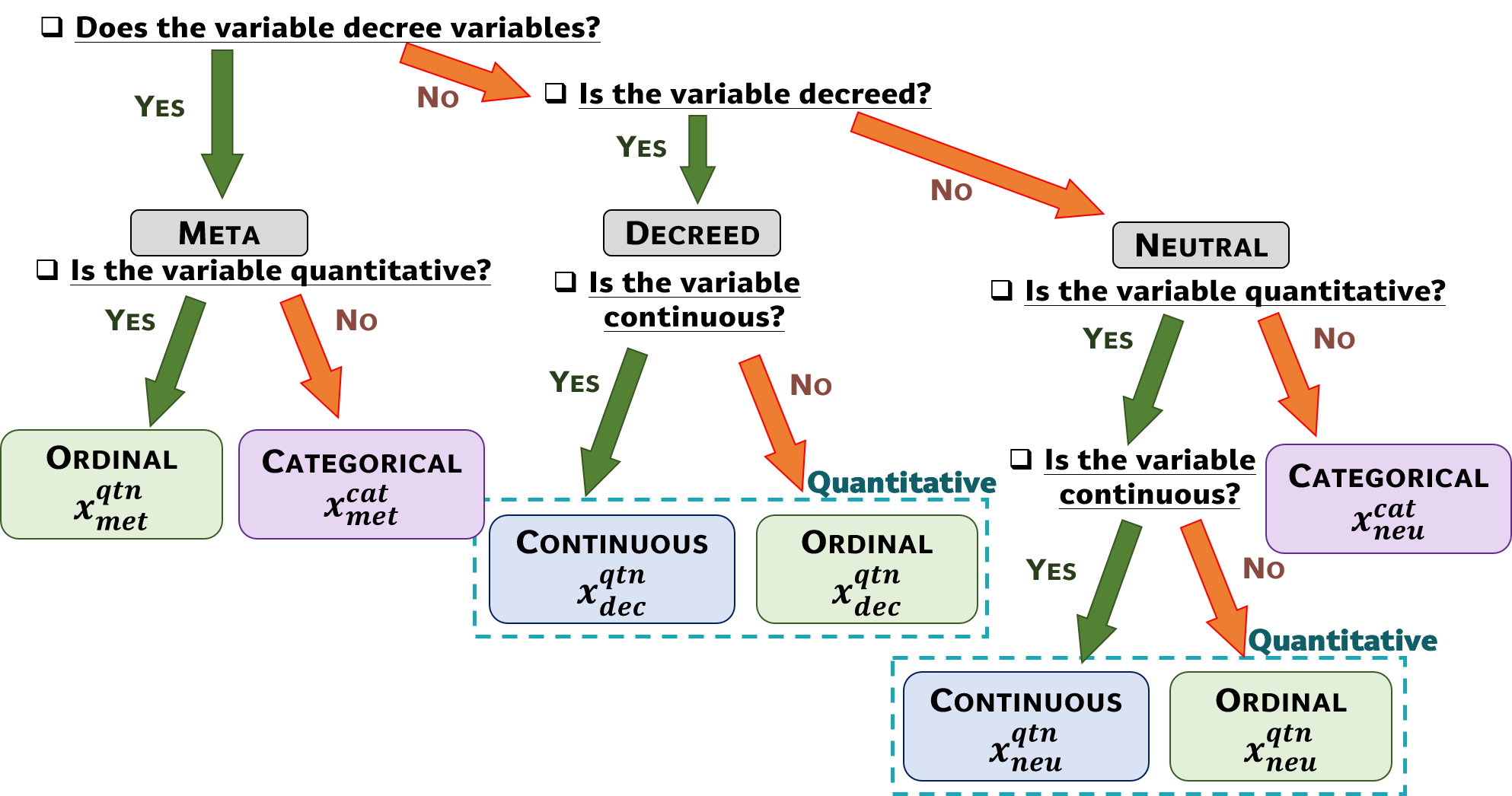}
\caption{Variables classification as used in \texttt{SMT 2.0}.
}
\label{fig:vars}
\end{figure}

To explain the framework and the new Kriging model, we illustrate the inputs variables of the model using a classical machine learning problem related to the hyperparameters optimization of a fully-connected Multi-Layer Perceptron (MLP)~\cite{audet2022general} on~\figref{fig:MLP}.
In~\tabref{tab:hyp_NN}, we detail the input variables of the model related to the MLP problem (i.e., the hyperparameters of the neural network, together with their types and roles). To keep things clear and concise, the chosen problem is a simplification of the original problem developed by \citet{audet2022general}.
Regarding the MLP problem of~\figref{fig:MLP} and following the classification scheme of~\figref{fig:vars}, we start by separating the input variables according to their role. In fact,
\begin{enumerate}
    \item changing the number of hidden layers modifies the number of inputs variables. Therefore, "\# of hidden layers" is a meta variable. 
    \item The number of neurons in the hidden layer number $k$ is either included or excluded. For example, the "\# of neurons in the $3^\text{rd}$ layer" would be excluded for an input that only has $2$ hidden layers. Therefore, "\# of neurons hidden layer $k$" are decreed variables.
    \item The "Learning rate", \textcolor{black}{"Momentum"}, "Activation function" and "Batch size" are not affected  by the hierarchy choice. Therefore, they are neutral variables. 
\end{enumerate}
According to their types, the MLP input variables can be classified as follows:
\begin{enumerate}[start=4]
    \item The meta variable "\# of hidden layers" is an integer and, as such, is represented by the component $x^{\quant}_{\meta}$. 
    \item The decreed variables "\# of neurons hidden layer $k$" are integers and, as such, are represented by the component $x^{\quant}_{\decreed}$. 
    \item The "Learning rate", \textcolor{black}{"Momentum"}, "Activation function" and "Batch size" are, respectively, continuous, \textcolor{black}{for the first two} (every value between two bounds), categorical (qualitative between \textcolor{black}{three} choices) and integer (quantitative between 6 choices). Therefore, the "Activation function" \textcolor{black}{and the "Momentum" are} represented by the component $x^{\cat}_{\neutral}$. The "Learning rate" and the "Batch size" are represented by the component $x^{\quant}_{\neutral}$.
\end{enumerate}
\begin{table}[H]
\centering 
\caption{A detailed description of the variables in the MLP problem.}
\resizebox{\linewidth}{!}{%
\begin{tabular}{@{}llclcc@{}}
\toprule
\multicolumn{2}{l}{\textbf{MLP Hyperparameters}}    & Variable &  Domain     & Type   & Role  \\ \midrule
\multicolumn{2}{l}{\textbf{Learning rate}}                & $r$      & $[10^{-5}, 10^{-2}]$       & FLOAT &  NEUTRAL  \\
\multicolumn{2}{l}{\textbf{\textcolor{black}{Momentum}}}                & $\textcolor{black}{\alpha}$      & $\textcolor{black}{[0, 1]}$       & \textcolor{black}{FLOAT} &  \textcolor{black}{NEUTRAL}  \\
\multicolumn{2}{l}{\textbf{Activation function}}          & $a$      & $ \{$ReLU, Sigmoid, \textcolor{black}{Tanh}$\}$         &  ENUM & NEUTRAL \\
\multicolumn{2}{l}{\textbf{Batch size}}                    & $b$        & $ \{8, 16, \ldots, 128, 256 \} $     & ORD      &  NEUTRAL  \\  
\multicolumn{2}{l}{\textbf{\# of hidden layers}}          & $l$      & $\{1,2,3\} $    &  ORD       & META       \\
\multicolumn{2}{l}{\textbf{\# of neurons hidden layer $k$}} & $n_{k}$ & $ \{50, 51 , \ldots, 55 \} $      & ORD & DECREED \\
\bottomrule
\end{tabular}
}
\label{tab:hyp_NN}
\end{table}
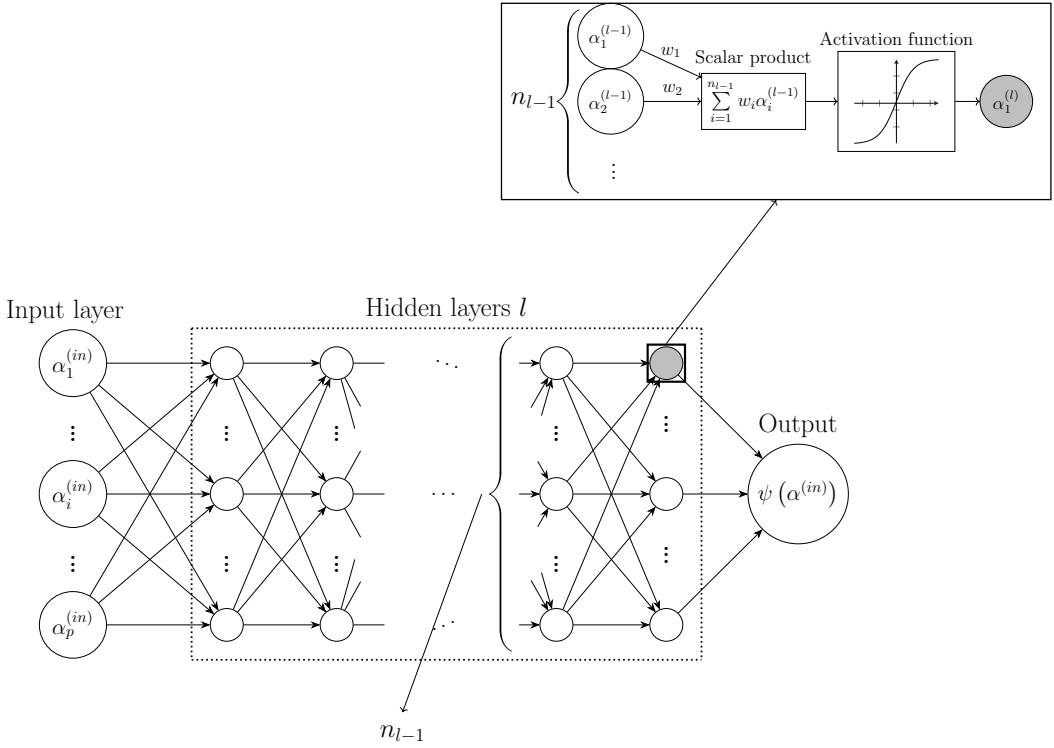
\begin{figure}[H]
\resizebox{\columnwidth}{!}{
\centering
\begin{tikzpicture}


\node [draw, circle, minimum size=0.75cm] (N1c) at (0,0) {\large $\alpha_i^{(in)}$};
\node [draw, circle, minimum size=0.75cm, xshift=0cm, yshift=3cm, at=(N1c), label={\hspace{-0.5cm} \vspace{0.5cm} \Large Input layer}] (N1u) {\large $\alpha_1^{(in)}$};
\node [draw, circle, minimum size=0.75cm, xshift=0cm, yshift=-3cm, at=(N1c)] (N1b) {\large $\alpha_p^{(in)}$};

\node [at=($(N1c)!0.5!(N1b)$)] {\Huge \vdots} ;
\node [at=($(N1c)!0.5!(N1u)$)] {\Huge \vdots} ;


\node [draw, circle, minimum size=0.75cm, xshift=3.5cm, yshift=0cm, at=(N1c)] (N2c) {};
\node [draw, circle, minimum size=0.75cm, xshift=0cm, yshift=3cm, at=(N2c)] (N2u) {};
\node [draw, circle, minimum size=0.75cm, xshift=0cm, yshift=-3cm, at=(N2c)] (N2b) {};

\node [at=($(N2c)!0.5!(N2b)$)] {\Huge \vdots} ;
\node [at=($(N2c)!0.5!(N2u)$)] {\Huge \vdots} ;

\node [draw, circle, minimum size=0.75cm, xshift=2.5cm, yshift=0cm, at=(N2c)] (N3c) {};
\node [draw, circle, minimum size=0.75cm, xshift=0cm, yshift=3cm, at=(N3c)] (N3u) {};
\node [draw, circle, minimum size=0.75cm, xshift=0cm, yshift=-3cm, at=(N3c)] (N3b) {};

\node [at=($(N3c)!0.5!(N3b)$)] {\Huge \vdots} ;
\node [at=($(N3c)!0.5!(N3u)$)] {\Huge \vdots} ;

\node [draw, circle, minimum size=0.75cm, xshift=5cm, yshift=0cm, at=(N3c)] (N6c) {}; 

\node [draw, circle, minimum size=0.75cm, xshift=0cm, yshift=3cm, at=(N6c)] (N6u) {};

\node [draw, circle, minimum size=0.75cm, xshift=0cm, yshift=-3cm, at=(N6c)] (N6b) {};

\node [at=($(N6c)!0.5!(N6b)$)] {\Huge \vdots} ;
\node [at=($(N6c)!0.5!(N6u)$)] {\Huge \vdots} ;

\node [draw, circle, minimum size=0.75cm, xshift=2.5cm, yshift=0cm, at=(N6c)] (N7c) {}; 
\node [draw, fill=black!25, circle, minimum size=0.75cm, xshift=0cm, yshift=3cm, at=(N7c)] (N7u) {};
\node [draw, circle, minimum size=0.75cm, xshift=0cm, yshift=-3cm, at=(N7c)] (N7b) {};

\node [at=($(N7c)!0.43!(N7b)$)] {\Huge \vdots} ;
\node [at=($(N7c)!0.57!(N7u)$)] {\Huge \vdots} ;


\node [draw, circle, minimum size=1cm, xshift=3cm, yshift=0cm, at=(N7c), label={ \Large Output}] (N8c) {\large $\psi \left(\alpha^{(in)}\right )$}; 

\draw[ -{Stealth[length=2mm, width=1.5mm]} ] (N7c)--(N8c);
\draw[ -{Stealth[length=2mm, width=1.5mm]} ] (N7u)--(N8c);
\draw[ -{Stealth[length=2mm, width=1.5mm]}] (N7b)--(N8c);

\node [ circle, minimum size=0.75cm, at=($(N3c)!.33!(N6c)$)] (N4c) {};
\node [ circle, minimum size=0.75cm, at=($(N3u)!.33!(N6u)$)] (N4u) {};
\node [ circle, minimum size=0.75cm, at=($(N3b)!.33!(N6b)$)] (N4b) {};

\node [ circle, minimum size=1cm, at=($(N3c)!.66!(N6c)$)] (N5c) {}; 
\node [ circle, minimum size=1cm, at=($(N3u)!.66!(N6u)$)] (N5u) {};
\node [ circle, minimum size=1cm, at=($(N3b)!.66!(N6b)$)] (N5b) {};


\draw[ -{Stealth[length=2mm, width=1.5mm]} ] (N1c)--(N2c);
\draw[ -{Stealth[length=2mm, width=1.5mm]} ] (N1c)--(N2u);
\draw[ -{Stealth[length=2mm, width=1.5mm]} ] (N1c)--(N2b);

\draw[ -{Stealth[length=2mm, width=1.5mm]} ] (N1u)--(N2c);
\draw[ -{Stealth[length=2mm, width=1.5mm]} ] (N1u)--(N2u);
\draw[ -{Stealth[length=2mm, width=1.5mm]} ] (N1u)--(N2b);

\draw[ -{Stealth[length=2mm, width=1.5mm]} ] (N1b)--(N2c);
\draw[ -{Stealth[length=2mm, width=1.5mm]} ] (N1b)--(N2u);
\draw[ -{Stealth[length=2mm, width=1.5mm]} ] (N1b)--(N2b);

\draw[ -{Stealth[length=2mm, width=1.5mm]} ] (N2c)--(N3c);
\draw[ -{Stealth[length=2mm, width=1.5mm]} ] (N2c)--(N3u);
\draw[ -{Stealth[length=2mm, width=1.5mm]} ] (N2c)--(N3b);

\draw[ -{Stealth[length=2mm, width=1.5mm]} ] (N2u)--(N3c);
\draw[ -{Stealth[length=2mm, width=1.5mm]} ] (N2u)--(N3u);
\draw[ -{Stealth[length=2mm, width=1.5mm]} ] (N2u)--(N3b);

\draw[ -{Stealth[length=2mm, width=1.5mm]} ] (N2b)--(N3c);
\draw[ -{Stealth[length=2mm, width=1.5mm]} ] (N2b)--(N3u);
\draw[ -{Stealth[length=2mm, width=1.5mm]} ] (N2b)--(N3b);

\draw[ - ] (N3c) -- ($(N3c)!.66!(N4c)$); 
\draw[ - ] (N3c) -- ($(N3c)!.33!(N4u)$);
\draw[ - ] (N3c) -- ($(N3c)!.33!(N4b)$);

\draw[ - ] (N3u)-- ($(N3u)!.33!(N4c)$);
\draw[ - ] (N3u)-- ($(N3u)!.66!(N4u)$); 
\draw[ - ] (N3u)-- ($(N3u)!.25!(N4b)$); 

\draw[ - ] (N3b)-- ($(N3b)!.33!(N4c)$);
\draw[ - ] (N3b)-- ($(N3b)!.25!(N4u)$); 
\draw[ - ] (N3b)-- ($(N3b)!.66!(N4b)$); 

\node [at=($(N4c)!0.5!(N5c)$)] {\large \dots } ;
\node [at=($(N4u)!0.5!(N5u)$), rotate=160] {\large \dots } ;
\node [at=($(N4b)!0.5!(N5b)$), rotate=20] {\large \dots } ;

\draw[ -{Stealth[length=2mm, width=1.5mm]}  ] ($(N5c)!.5!(N6c)$)--(N6c); 
\draw[ -{Stealth[length=2mm, width=1.5mm]}  ] ($(N5c)!.66!(N6u)$)--(N6u);
\draw[ -{Stealth[length=2mm, width=1.5mm]}  ] ($(N5c)!.66!(N6b)$)--(N6b);

\draw[ -{Stealth[length=2mm, width=1.5mm]}  ] ($(N5u)!.75!(N6c)$)--(N6c);
\draw[ -{Stealth[length=2mm, width=1.5mm]}  ] ($(N5u)!.5!(N6u)$)--(N6u); 
\draw[ -{Stealth[length=2mm, width=1.5mm]}  ] ($(N5u)!.8!(N6b)$)--(N6b); 

\draw[ -{Stealth[length=2mm, width=1.5mm]}  ] ($(N5b)!.75!(N6c)$)--(N6c);
\draw[ -{Stealth[length=2mm, width=1.5mm]}  ] ($(N5b)!.8!(N6u)$)--(N6u); 
\draw[ -{Stealth[length=2mm, width=1.5mm]}  ] ($(N5b)!.5!(N6b)$)--(N6b); 

\draw[ -{Stealth[length=2mm, width=1.5mm]} ] (N6c)--(N7c);
\draw[ -{Stealth[length=2mm, width=1.5mm]} ] (N6c)--(N7u);
\draw[ -{Stealth[length=2mm, width=1.5mm]} ] (N6c)--(N7b);

\draw[ -{Stealth[length=2mm, width=1.5mm]} ] (N6u)--(N7c);
\draw[ -{Stealth[length=2mm, width=1.5mm]} ] (N6u)--(N7u);
\draw[ -{Stealth[length=2mm, width=1.5mm]} ] (N6u)--(N7b);

\draw[ -{Stealth[length=2mm, width=1.5mm]} ] (N6b)--(N7c);
\draw[ -{Stealth[length=2mm, width=1.5mm]} ] (N6b)--(N7u);
\draw[ -{Stealth[length=2mm, width=1.5mm]} ] (N6b)--(N7b);

\node [circle, minimum size=1cm] (T1u) at ($(N2u)-(0.8,-0.8)$)  {};
\node [circle, minimum size=1cm] (T1b) at ($(N2b)-(0.8,0.8)$)  {};
\node [circle, minimum size=1cm] (T2u) at ($(N7u)+(0.8,0.8)$)  {};
\node [circle, minimum size=1cm] (T2b) at ($(N7b)+(0.8,-0.8)$)  {};
\draw[dotted, line width= 0.4mm] ($(T1u)+(0,0)$)--($(T1b)+(0,0)$);
\draw[dotted, line width= 0.4mm] ($(T2u)+(0,0)$)--($(T2b)+(0,0)$);

\draw[dotted, line width= 0.4mm] ($(T1u)+(0,0)$)--($(T2u)+(0,0)$) node[midway, above, black, 
]{\Large Hidden layers $l$} ;

\draw[dotted, line width= 0.4mm] ($(T1b)+(0,0)$)--($(T2b)+(0,0)$);

\draw [decorate, decoration = {calligraphic brace,raise=15pt, mirror, amplitude=14pt}, line width = 0.4mm ] ($(N6u)-(0.5,-0.6)$) --  ($(N6b)-(0.5,0.6)$); 

\draw[->] ($(N5c)+(0,0)$) -- (7.5, -5) node[inner sep=8pt, at end, yshift=-0.5cm] {\Large $n_{l-1}$} ;

\node [draw, minimum size=0.85cm, line width=0.5mm, at=(N7u)] (N7U) {};

\node [draw, minimum height=4.5cm, minimum width=12.5cm, xshift=2.5cm, yshift=6cm, line width=0.3mm, at=(N7U)] (box) {}; 

\draw[->, inner sep=5pt] (N7U.north) .. controls +(down:0cm) and +(right:0cm) .. (box.south);

\node (boxC) [draw, circle, minimum size=0.85cm, at=(box.west), xshift=2.5cm] {$\alpha_{2}^{(l-1)}$}  ;

\node (boxCC) [ minimum size=0.85cm, at=(box.west), xshift=0.75cm] {\Large$n_{l-1}$}  ;

\node (boxT) [draw, circle, minimum size=0.85cm, at=(boxC), yshift=1.5cm] {$\alpha_{1}^{(l-1)}$}  ;
\node (boxB) [ circle, minimum size=0.85cm, at=(boxC), yshift=-1.5cm] { \vspace{-1cm} \Large \vdots }  ;
\draw [decorate, decoration = {calligraphic brace,raise=15pt, mirror, amplitude=14pt}, line width = 0.4mm ] ($(boxT)-(0.2,-0.6)$) --  ($(boxB)-(0.2,0.6)$) ;


\node (boxScal) [draw, minimum size=0.85cm, at=(boxC), xshift=3.25cm, label={Scalar product}] { 
$   \sum \limits_{i=1}^{n_{l-1}} w_i \alpha_i^{(l-1)} $
}  ;

\draw[->] (boxT) -- (boxScal) node[midway, above, black]{$w_{1}$};
\draw[->] (boxC) -- (boxScal) node[midway, above, black]{$w_{2}$};

\node (boxAct) [draw, minimum size=0.85cm, at=(boxScal), xshift=3.25cm, label={Activation function}] {
\begin{tikzpicture}
        \begin{axis}[axis lines=center, width=3.5cm, axis x line shift=1cm, height=3.5cm,
         yticklabels={,,},
         xticklabels={,,}
         ]
        \addplot[color=black]{1/(1+exp(-x))-2};
        \end{axis}
        \end{tikzpicture}
}  ;

\draw [->] (boxScal) -- (boxAct);

\node (nodeRed) [draw, circle, minimum size=0.85cm, fill=black!25, at=(boxAct), xshift=2.5cm] {$\alpha_{1}^{(l)}$};

\draw [->] (boxAct) -- (nodeRed);

\end{tikzpicture}}
\caption{The Multi-Layer Perceptron (MLP) problem (figure adapted from~\cite[Figure 1]{audet2022general}).}
\label{fig:MLP} 
\end{figure}

To model hierarchical variables, as proposed in~\cite{audet2022general}, we separate the input space $\mathcal{X}$ as $( \mathcal{X}_{\neutral},\mathcal{X}_{\meta},\mathcal{X}_{\decreed})$ where $\displaystyle \mathcal{X}_{\decreed}= \bigcup_{x_\meta \in \mathcal{X}_{\meta}} \mathcal{X}_{\acting}(x_\meta) $. 
Hence, for a given point $x \in \mathcal{X}$, one has $x=(x_{\neutral}, x_{\meta}, x_{\acting} (x_{\meta}) ),$ where $x_{\neutral} \in \mathcal{X}_{\neutral}$, $x_{\meta}\in \mathcal{X}_{\meta}$ and $x_{\acting} (u_{\meta})\in \mathcal{X}_{\acting} (u_{\meta})$ are defined as follows:
 \begin{itemize}
     \item The components $x_{\neutral} \in \mathcal{X}_{\neutral}$ gather all neutral variables that are not impacted by the meta variables but needed. 
     For example, in the MLP problem, $\mathcal{X}_{\neutral}$ gathers the possible learning rates, \textcolor{black}{momentum}, activation functions and batch sizes.
     Namely, from \textcolor{black}{Table}~\ref{tab:hyp_NN}, $ \mathcal{X}_{\neutral } = [10^{-5},10^{-2}] \times \textcolor{black}{ [0,1] \times \{\mbox{ReLu}, \mbox{Sigmoid},\mbox{Tanh} \}} \times \{8,16,\ldots,256\} $.

     \item The components $x_{\meta}$ gather the meta (also known as dimensional) variables that determine the inclusion or exclusion of other variables. 
     For example, in the MLP problem, $\mathcal{X}_{\meta}$ represents the possible numbers of layers in the MLP. 
     Namely, from \textcolor{black}{Table}~\ref{tab:hyp_NN}, $ \mathcal{X}_{\meta} = \{1,2,3\}$.


     \item The components $x_{\acting} (x_{\meta})$, contain the decreed variables whose inclusion (decreed-included) or exclusion (decreed-excluded) is determined by the values of the meta components $x_{\meta}$. 
     For example, in the MLP problem, $\mathcal{X}_{\decreed}$ represents the number of neurons in the decreed layers.
     Namely, from \textcolor{black}{Table}~\ref{tab:hyp_NN}, $ \mathcal{X}_{\acting} (x_\meta=3) =  [50,55]^3$, $ \mathcal{X}_{\acting} (x_\meta=2) =  [50,55]^2$ and $ \mathcal{X}_{\acting} (x_\meta=1) =  [50,55]$. 
 \end{itemize}


\subsection{A Kriging model for hierarchical variables}

In this section, a new method to build a Kriging model with hierarchical variables is introduced based on the framework aforementioned. 
The proposed methods are included in \texttt{SMT 2.0}.
\subsubsection{Motivation and state-of-the-art}

Assuming that the decreed variables are quantitative,~\citet{Hutter} proposed several kernels for the hierarchical context. 
A classic approach, called the imputation method (\texttt{Imp-Kernel}) leads to an efficient paradigm in practice that can be easily integrated into a more general framework as proposed by~\citet{Effectiveness}. 
However, this kernel lacks depth and depends on arbitrary choices.
Therefore,~\citet{Hutter} also proposed a more general kernel, called \texttt{Arc-Kernel} and~\citet{Horn_hier} generalized this kernel even more and proposed a new formulation called the \texttt{Wedge-Kernel}~\cite{DACE_hier}.   
The drawbacks of these two methods are that they add some extra hyperparameters for every decreed dimension (respectively one extra hyperparameter for the \texttt{Arc-Kernel} and two hyperparameters for the \texttt{Wedge-Kernel}) and that they need a normalization according to the bounds. 
More recently,~\citet{pelamattihier} developed a hierarchical kernel for Bayesian optimization. 
However, our work is also more general thanks to the framework introduced earlier~\cite{audet2022general} that considers variable-wise formulation and is more flexible as we are allowing sub-problems to be intersecting. 

In the following, we describe our new method to build a correlation kernel for hierarchical variables.
In particular, we introduce a new algebraic kernel called \texttt{Alg-Kernel} that behaves like the \texttt{Arc-Kernel} whilst correcting most of its drawbacks. 
In particular, our kernel does not add any hyperparameters, and the normalization is handled in a natural way.

\subsubsection{A new hierarchical correlation kernel} 

For modeling purposes, we assume that the decreed space is quantitative, i.e., $ \mathcal{X}_{\decreed}= \mathcal{X}_{\decreed}^{\quant}$.
Let  $u \in \mathcal{X}$ be an input point partitioned as $u=(u_{\neutral}, u_\meta, u_{\acting}(u_{\meta}))$  and, similarly, $v\in \mathcal{X}$ is another input such that $v=(v_{\neutral}, v_\meta, v_{\acting}(v_{\meta}))$. 
The new kernel $k$ that we propose for hierarchical variables is given by
\begin{eqnarray} 
k(u, v) &= &k_{\neutral} (u_{\neutral},v_{\neutral}) \times  
k_\meta(u_\meta,v_\meta) \nonumber \\
& & \quad \quad \times \ 
k_{\meta,\decreed} ( [u_\meta, u_{\acting}(u_\meta)], [v_\meta, v_{\acting}(v_\meta)]), 
\label{eq:hier_ker}
\end{eqnarray}
where $k_{\neutral}$, $k_\meta$ and $k_{\meta,\decreed}$ are as follows:
 \begin{itemize}
\item  $k_{\neutral}$ represents the neutral kernel that encompasses both categorical and quantitative neutral variables, i.e., $k_{\neutral}$ can be decomposed into two parts $k_{\neutral} (u_{\neutral},v_{\neutral})= k^{\cat}(u_{\neutral}^{\cat},v_{\neutral}^{\cat})k^{\quant} (u_{\neutral}^{\quant},v_{\neutral}^{\quant}).$
The categorical kernel, denoted $k^{\cat}$, could be any Symmetric Positive Definite (SPD)~\cite{Mixed_Paul} mixed kernel (see Section~\ref{sec:mixed}). 
For the quantitative (integer or continuous) variables, a distance-based kernel is used. 
The  chosen quantitative kernel (Exponential, Matérn,...), always depends on a given distance $d$.
For example, the $n$-dimensional exponential kernel gives 

\begin{equation}  
 k^{\quant}(u^{\quant},v^{\quant}) = \displaystyle{ \prod^{n}_{i=1} \exp ( - d(u^{\quant}_i,v^{\quant}_i))}.
\end{equation}

    \item $k_{\meta}$ is the meta variables related kernel.
    It is also separated into two parts:
$k_{\meta} (u_{\meta},v_{\meta})= k^{\cat}(u_{\meta}^{\cat},v_{\meta}^{\cat})k^{\quant} (u_{\meta}^{\quant},v_{\meta}^{\quant})$ where the quantitative kernel is ordered and not continuous because meta variables take value in a finite set. 
\item  $k_{\meta,\decreed}$ is  an SPD kernel that models the correlations between the meta levels (all the possible subspaces) and the decreed variables. 
In what comes next, we detailed this kernel.
\end{itemize}

\subsubsection{Towards an algebraic meta-decreed kernel} 
Meta-decreed kernels like the imputation kernel or the \texttt{Arc-Kernel} were first proposed in~\cite{Zaefferer,Hutter} and the distance-based kernels such as \texttt{Arc-Kernel} or \texttt{Wedge-Kernel}~\cite{DACE_hier} were proven to be SPD. 
Nevertheless, to guarantee this SPD property, the same hyperparameters are used to model the correlations between the meta levels and between the decreed variables~\cite{Zaefferer}. 
For this reason, the \texttt{Arc-Kernel} includes additional continuous hyperparameters which makes the training of the GP models more expensive and introduces more numerical stability issues. 
In this context, we are proposing a new  algebraic meta-decreed kernel (denoted as \texttt{Alg-Kernel}) that enjoys similar properties as \texttt{Arc-Kernel} but without using additional continuous hyperparameters nor rescaling. 
In the \texttt{SMT 2.0} release, we included \texttt{Alg-Kernel} and a simpler version of \texttt{Arc-Kernel} that do not relies on additional hyperparameters.

Our proposed \texttt{Alg-Kernel} kernel is given by 
\begin{equation}
\begin{split}
 & k^{\text{alg}}_{\meta,\decreed} ( [u_\meta, u_{\acting}(u_\meta)], [v_\meta, v_{\acting}(v_\meta)]) \\&  \quad \quad = k^{\text{alg}}_\meta(u_\meta, v_\meta) \times   k^{\text{alg}}_{\decreed}(u_{\acting}(u_\meta),v_{\acting}(v_\meta)).
\end{split}
\end{equation}
Mathematically, we could consider that there is only one meta variable whose levels correspond to every possible included subspace. 
Let $I_{\text{sub}}$ denotes the components indices of possible subspaces, the subspaces parameterized by the meta component $u_\meta$ are defined as $\mathcal{X}_{\acting}(u_\meta=l), \ l \in I_{\text{sub}} $.
It follows that the fully extended continuous decreed space writes as $\mathcal{X}_{\decreed} = \bigcup_{l \in I_{\text{sub}}} \mathcal{X}_{\acting}(u_\meta=l)$ and $I_\decreed$ is the set of the associated indices. 
Let $I^{inter}_{u,v}$ denotes the set of components related to the space $ \mathcal{X}_{\acting} (u_\meta,v_\meta)$ containing the variables decreed-included in both  $ \mathcal{X}_{\acting} (u_\meta)$ and  $ \mathcal{X}_{\acting} (v_\meta) $.

Since the decreed variables are quantitative, one has 
\begin{equation}
\begin{split}
k^{\text{alg}}_{\decreed} (u_{\acting}(u_\meta),v_{\acting}(v_\meta)) 
&= k^{\quant} (u_{\acting}(u_\meta),v_{\acting}(v_\meta))\\
& = \prod_{i \in I_{u,v}^{inter}}   k^{\quant} (  [(u_{\acting}(u_\meta)]_i,[v_{\acting}(v_\meta)]_i )
\end{split}
\end{equation}
The construction of the quantitative kernel $k^{\quant}$ depends on a given distance denoted $d^{\text{alg}}$. 
The kernel $k^{\text{alg}}_\meta$ is an induced meta kernel that depends on the same distance $d^{\text{alg}}$ to preserve the SPD property of $k^{\text{alg}}_{\meta,\decreed}$. 
For every $i \in I_\decreed$, if $i \in I^{inter}_{u,v} $, the new algebraic distance is given by
\begin{equation}
d^{\text{alg}}( [u_{\acting} (u_\meta) ]_i , [v_{\acting} (v_\meta) ]_i )  = \left(\frac{2  | [u_{\acting} (u_\meta) ]_i  - [v_{\acting} (v_\meta) ]_i|}{ \sqrt{{ [u_{\acting} (u_\meta) ]_i }^2+1}\sqrt{{ [v_{\acting} (v_\meta) ]_i }^2+1}}\right)\theta_i,
\end{equation}
where $\theta_i \in \mathbb{R}^+$ is a continuous hyperparameter.
Otherwise, if $i \in I_\decreed$ but $i \notin I^{inter}_{u,v} $, there should be a non-zero residual distance between the two different subspaces $ \mathcal{X}_{\acting} (u_\meta)$ and $\mathcal{X}_{\acting} (v_\meta)$ to ensure the kernel SPD property. 
To have a residual not depending on the decreed values, our model considers that there is a unit distance
$$d^{\text{alg}}( [u_{\acting} (u_\meta) ]_i , [v_{\acting} (v_\meta) ]_i) = 1.0 \ \theta_i, \  \forall i \in I_\decreed \setminus I^{inter}_{u,v}. $$ 
%
The induced meta kernel $k^{\text{alg}}_{\meta}(u_{\meta},v_{\meta})$ to preserve the SPD property of $k^{\text{alg}}$ is defined as: 
\begin{equation}
k^{\text{alg}}_{\meta}(u_{\meta},v_{\meta}) 
=   \prod_{i \in I_\meta} \ k^{\quant}(1.0 \ \theta_i) .  
\label{eq:d_alg}
\end{equation}
%
Not only our kernel of~\eqnref{eq:hier_ker} uses less hyperparameters than the \texttt{Arc-Kernel} (as we cut off its extra parameters) but it is also a more flexible kernel as it allows different kernels for meta and decreed variables.
Moreover, another advantage of our kernel is that it is numerically more stable thanks to the new non-stationary~\cite{hebbal2021bayesian} algebraic distance defined in~\eqnref{eq:d_alg}~\cite{wildberger2007rational}. 
Our proposed distance also does not need any rescaling by the bounds to have values between 0 and 1.

In what comes next, we will refer to the implementation of the kernels \texttt{Arc-Kernel} and \texttt{Alg-Kernel} by \texttt{SMT Arc-Kernel} and \texttt{SMT Alg-Kernel}. 
We note also that the implementation of \texttt{SMT Arc-Kernel} differs slightly from the original \texttt{Arc-Kernel} as we fixed some hyperparameters to 1 in order to avoid adding extra hyperparameters and use the formulation of~\eqnref{eq:hier_ker} and rescaling of the data. 

\subsubsection{Illustration on the MLP problem}

In this section, we illustrate the hierarchical \texttt{Arc-Kernel} on the MLP example. For that sake, we consider two design variables $u$ and $v$ such that  $u = (2.10^{-4}, \textcolor{black}{0.9}, \mbox{ReLU}, 16, 2, 55,51)$ and $v =  ( 5.10^{-3},\textcolor{black}{0.8}, \mbox{Sigmoid}, 64, 3, 50,54,53)$. 
Since the value of $u_\meta$ (i.e., the number of hidden layers) differs from one point to another (namely, $2$ for $u$  and $3$ for $v$), the associated variables $u_\acting(u_\meta)$ have  either 2 or 3 variables for the number of neurons in each layer (namely 55 and 51 for $u$, and 50, 54 and 53 for the point $v$).
In our case, \textcolor{black}{8} hyperparameters $([R_1]_{1,2}, \theta_1, \ldots, \textcolor{black}{\theta_7})$ will have to be optimized where $k$ is given by \eqnref{eq:hier_ker}.
These 7 hyperparameters can be described using our proposed framework as follows:
\begin{itemize}
    \item For the neutral components, we have $u_\neutral = (2.10^{-4}, \textcolor{black}{0.9} ,\mbox{ReLU}, 16)$ and $v_\neutral = (5.10^{-3},\textcolor{black}{0.8}, \mbox{Sigmoid}, 64)$. Therefore, for a categorical matrix kernel $R_1$  and a square exponential quantitative kernel,
\begin{equation*}
\centering
\begin{split}
k_{\neutral} (u_{\neutral},v_{\neutral}) &=  k^{\cat}(u_{\neutral}^{\cat},v_{\neutral}^{\cat})k^{\quant} (u_{\neutral}^{\quant},v_{\neutral}^{\quant}) \\ 
&= [R_1]_{1,2} \exp{[- \theta_1 (2.10^{-4} -5.10^{-3})^2]  }  \\
&\quad \ \ \textcolor{black}{ \exp{[- \theta_2 (0.9 - 0.8)^2]  } }  \exp{[ - \textcolor{black}{\theta_3} (16-64)^2 ]}.
\end{split}
\end{equation*}
The values $[R_1]_{1,2}$, $\theta_1$, \textcolor{black}{$\theta_2$} and $\theta_3$ need to be optimized. Here, $[R_1]_{1,2}$ is the correlation between "ReLU" and "Sigmoid". 

\item 
For the meta components, we have $u_\meta = 2$ and $v_\meta = 3$.
Therefore, for a square exponential quantitative kernel,
\begin{equation*}
\centering
\begin{split}
k_{\meta} (u_{\meta},v_{\meta})&= k^{\cat}(u_{\meta}^{\cat},v_{\meta}^{\cat})k^{\quant} (u_{\meta}^{\quant},v_{\meta}^{\quant}) \\
&= \exp{[ - \textcolor{black}{\theta_4} (3-2)^2 ]}.
\end{split}
\end{equation*}
The value \textcolor{black}{$\theta_4$} needs to be optimized. 

\item  For the meta-decreed kernel, we have $[ u_\meta, u_\acting(u_\meta) ] = [2, (55,51)] $ and $[ v_\meta, v_\acting(v_\meta) ] = [3, (50,54,53)]$ which gives
\begin{eqnarray*}
 & & k^{\text{alg}}_{\meta,\decreed} ( [u_\meta, u_{\acting}(u_\meta)], [v_\meta, v_{\acting}(v_\meta)]) \\& &  \quad \quad = k^{\mathrm{alg}}_\meta(2,3) \ k^{\mathrm{alg}}_\decreed( (55,51), (50,54,53)).
\end{eqnarray*}
The distance $d^\text{alg} (51, 54) =  \left( \frac{  2 \times |51-54|}{ \sqrt{ 51^2+1 } \sqrt{ 54^2+1 }} \right) \textcolor{black}{\theta_6} = 2.178.10^{-3} \ \textcolor{black}{\theta_6}.$
In general, for surrogate models, and in particular in \texttt{SMT 2.0}, the input data are normalized.
With a unit normalization from $[50,55]$ to $[0,1]$, we would have  $d^\text{alg} (0.2, 0.8) =  \left( \frac{ 2 \times 0.6}{ \sqrt{ 0.2^2+1 } \sqrt{ 0.6^2+1 }} \right) \textcolor{black}{\theta_6} = 0.919 \ \textcolor{black}{\theta_6} .$ 
Similarly, we have, between 55 and 50, $d^\text{alg} (0, 1) =  1.414 \ \textcolor{black}{\theta_5}.$ 
Hence, for a square exponential quantitative kernel, one gets
\begin{eqnarray*}
 & & k^{\text{alg}}_{\meta,\decreed} ( [u_\meta, u_{\acting}(u_\meta)], [v_\meta, v_{\acting}(v_\meta)]) \\
& &  \quad \quad = \exp{[ - \textcolor{black}{\theta_7}]} \times \exp{[ - 1.414\ \textcolor{black}{\theta_5} ]} \times \exp{[ - 0.919\ \textcolor{black}{\theta_6} ]},
\end{eqnarray*} 
where the meta induced component is $k^{\text{alg}}_{\meta}(u_{\meta},v_{\meta}) =  \exp{[ -{\theta_7}] }$ because the decreed value $53$ in $v$ has nothing to be compared with in $u$ as in~\eqnref{eq:d_alg}.
The values  \textcolor{black}{$\theta_5$, $\theta_6$ and $\theta_7$} need to be optimized which complete the description of the hyperparameters.

We note that for the MLP problem, \texttt{Alg-Kernel} models use \textcolor{black}{10} hyperparameters whereas the \texttt{Arc-Kernel} would require \textcolor{black}{12} hyperparameters without the meta kernel ($\textcolor{black}{\theta_4}$) but with 3 extra decreed hyperparameters 
and the \texttt{Wedge-Kernel} would require \textcolor{black}{15} hyperparameters.
For deep learning applications, a more complex perceptron with up to $10$ hidden layers would require $\textcolor{black}{17}$ hyperparameters with \texttt{SMT 2.0} models against $\textcolor{black}{26}$ for \texttt{Arc-Kernel} and $\textcolor{black}{36}$ for \texttt{Wedge-Kernel}. 
The next section illustrates the interest of our method to build a surrogate model for this neural network engineering problem.

\end{itemize}

\subsection{A neural network test-case using \texttt{SMT 2.0}}
\label{sec:NN}
In this section, we apply our models to the hyperparameters optimization of a MLP problem aforementioned of~\figref{fig:MLP}. 
Within \texttt{SMT 2.0} an example illustrates this MLP problem. 
For the sake of showing the Kriging surrogate abilities, we implemented a dummy function with no significance to replace the real black-box that would require training a whole Neural Network (NN) with big data. 
This function requires a number of variables that depends on the value of the meta variable, i.e the number of hidden layers. 
To simplify, we have chosen only 1, 2 or 3 hidden layers and therefore, we have 3 decreed variables but deep neural networks could also be investigated as our model can tackle a few dozen variables. 
A test case (\emph{test\_hierarchical\_variables\_NN}) shows that our model \texttt{SMT Alg-Kernel} interpolates the data properly, checks that the data dimension is correct and also asserts that the inactive decreed variables have no influence over the prediction. 
In~\figref{fig:NN_hier} we illustrate the usage of Kriging surrogates with hierarchical and mixed variables based on the implementation of \texttt{SMT 2.0} for \emph{test\_hierarchical\_variables\_NN}. 

To compare the hierarchical models of \texttt{SMT 2.0} (\texttt{SMT Alg-Kernel} and \texttt{SMT Arc-Kernel}) with the state-of-the-art imputation method previously used on industrial application (\texttt{Imp-Kernel})~\cite{Effectiveness}, we draw a 99 point LHS (33 points by meta level) as a training set and the validation set is a LHS of $3\times 1000=3000$ points. 
For the \texttt{Imp-Kernel}, the decreed-excluded values are replaced by $52$ because the mean value $52.5$ is not an integer (by default, \texttt{SMT} rounds to the floor value). 
For the three methods, the precision (computed with a root-mean-square error RMSE criterion), the likelihood and the computational time are shown in~\tabref{tab:resNN}.
For this problem, we can see that \texttt{SMT Alg-kernel} gives better performance than the imputation method or \texttt{SMT Arc-kernel}. Also, as all methods use the same number of hyperparameters, they have similar time performances. 
A direct application of our modeling method is Bayesian optimization to perform quickly the hyperparameter optimization of a neural network~\cite{cho2020basic}. 

\begin{table}[H]
\centering
\caption{Results on the neural network model. }
\small
\resizebox{\columnwidth}{!}{
\begin{tabular}{cccc} 
\hline
\textbf{Hierarchical method}  &  Prediction error (RMSE) & $ \ $ Likelihood  & \# of hyperparam. 
\\
\hline 
\texttt{SMT Alg-kernel}  & \ {\textcolor{black}{3.7610}
} & {
\textcolor{black}{176.11}
} & \textcolor{black}{10}
\\   
\texttt{SMT Arc-kernel}  & \textcolor{black}{4.9208} 
& \textcolor{black}{162.01}
& \textcolor{black}{10} 
\\   
\texttt{Imp-Kernel}  & \textcolor{black}{4.5455}
& \textcolor{black}{170.64}
& \textcolor{black}{10} 
\\
\hline
\end{tabular}
}
\label{tab:resNN}
\end{table}
\begin{figure}
\vspace{-65pt}
\begin{lstlisting}
from smt.sampling_methods import LHS
from smt.surrogate_models import KRG,MixIntKernelType,\
  MixHrcKernelType,DesignSpace,FloatVariable,\
  IntegerVariable,OrdinalVariable,CategoricalVariable
from smt.applications.mixed_integer import \
  MixedIntegerSamplingMethod,MixedIntegerKrigingModel
import f1_NN, f2_NN,f3_NN #dummy example
def test_hierarchical_variables_NN(self):
   def dummy_f(x):
       if x[0] == 1:
         y=(f1_NN(x[1],x[2],x[3],2**x[4],x[5]))
       elif x[0] == 2:
         y=(f2_NN(x[1],x[2],x[3],2**x[4],x[5],x[6]))
       elif x[0] == 3:
         y=(f3_NN(x[1],x[2],x[3],2**x[4],x[5],x[6],x[7]))
       return y
   # Define the mixed hierarchical design space
   ds = DesignSpace([
     IntegerVariable(1, 3), FloatVariable(1e-5, 1e-2),
     FloatVariable(0, 1),
     CategoricalVariable(["ReLU", "Sigmoid","Tanh"]),
     IntegerVariable(3, 8), IntegerVariable(50, 55),
     IntegerVariable(50, 55), IntegerVariable(50, 55),
   ])
   # activate x5 when x0 in [2, 3]; x6 when x0 == 3
   ds.declare_decreed_var(
     decreed_var=6, meta_var=0, meta_value=[2, 3])
   ds.declare_decreed_var(7, meta_var=0, meta_value=3)
   #Perform the mixed integer sampling 
   sampling = MixedIntegerSamplingMethod(
    LHS, ds, criterion="ese", random_state=42)
   Xt = sampling(100)
   Yt = dummy_f(Xt)
   #Build the surrogate
   sm = MixedIntegerKrigingModel(
     surrogate=KRG(design_space=ds, corr="abs_exp",
     categorical_kernel=MixIntKernelType.HOMO_HSPHERE,
     hierarchical_kernel=MixHrcKernelType.ALG_KERNEL)
   sm.set_training_values(Xt, Yt)
   sm.train()
   # Check prediction accuracy
   y_s = sm.predict_values(Xt)
   pred_RMSE = np.linalg.norm(y_s - Yt) / len(Yt)
   y_sv = sm.predict_variances(Xt)
   var_RMSE = np.linalg.norm(y_sv) / len(Yt)
   assert pred_RMSE < 1e-7
   assert var_RMSE < 1e-7
\end{lstlisting}
\caption{Example of usage of Hierarchical and Mixed Kriging surrogate\textcolor{black}{.}}
\label{fig:NN_hier}
\end{figure}

\section{Bayesian optimization within \texttt{SMT 2.0}}
\label{sec:BO}

Efficient global optimization (EGO) is a sequential Bayesian optimization algorithm designed to find the optimum of a black-box function that may be expensive to evaluate~\cite{Jones98}.
EGO starts by fitting a Kriging model to an initial DoE, and then uses an acquisition function to select the next point to evaluate.
The most used acquisition function is the expected improvement. 
Once a new point has been evaluated, the Kriging model is updated.
Successive updates increase the model accuracy over iterations.
This enrichment process repeats until a stopping criterion is met.

Because \texttt{SMT 2.0} implements Kriging models that handle mixed and hierarchical variables, we can use EGO to solve problems involving such design variables. 
Other Bayesian optimization algorithms often adopt approaches based on solving subproblems with continuous or non-hierarchical Kriging.
This subproblem approach is less efficient and scales poorly, but it can only solve simple problems. 
Several Bayesian optimization software packages can handle mixed or hierarchical variables with such a subproblem approach.
The packages include BoTorch~\cite{balandat2020botorch}, SMAC~\cite{SMAC3}, Trieste~\cite{picheny2023trieste}, HEBO~\cite{cowen-rivers_hebo_2020}, OpenBox~\cite{jiang2023openbox}, and Dragonfly~\cite{Dragonfly}. 

\subsection{A mixed optimization problem} 
\label{sec:MI-BO}

\figref{res_optim_mi} compares the four EGO methods implemented in \texttt{SMT 2.0}: \texttt{SMT GD}, \texttt{SMT CR}, \texttt{SMT EHH} and \texttt{SMT HH}. 
The mixed test case that illustrates Bayesian optimization is a toy test case~\cite{CAT-EGO} detailed in~\ref{app:Toy}. 
This test case has two variables, one continuous and one categorical with 10 levels.
To assess the performance of our algorithm, we performed 20 runs with different initial DoE sampled by LHS.
Every DoE consists of 5 points and we chose a budget of 55 infill points.
\figref{convmi} plots the convergence curves for the four methods. 
In particular, the median is the solid line, and the first and third quantiles are plotted in dotted lines. 
To visualize better the data dispersion, the boxplots of the 20 best solutions after 20 evaluations are plotted in~\figref{mini_mi}. 
As expected, the more a method is complex, the better the optimization. Both \texttt{SMT HH} and \texttt{SMT EHH} converged in around 18 evaluations whereas \texttt{SMT CR} and \texttt{SMT GD} take around 26 iterations as shown on~\figref{convmi}.
Also, the more complex the model, the closer the optimum is to the real value as shown on~\figref{mini_mi}. 

\begin{figure}[H]
\begin{minipage}[b]{.6\linewidth}
\centering
\hspace{-1.25cm}
\subfloat[Convergence curves: medians of 20 runs.]{
\includegraphics[height=5cm,,width=7cm]
{  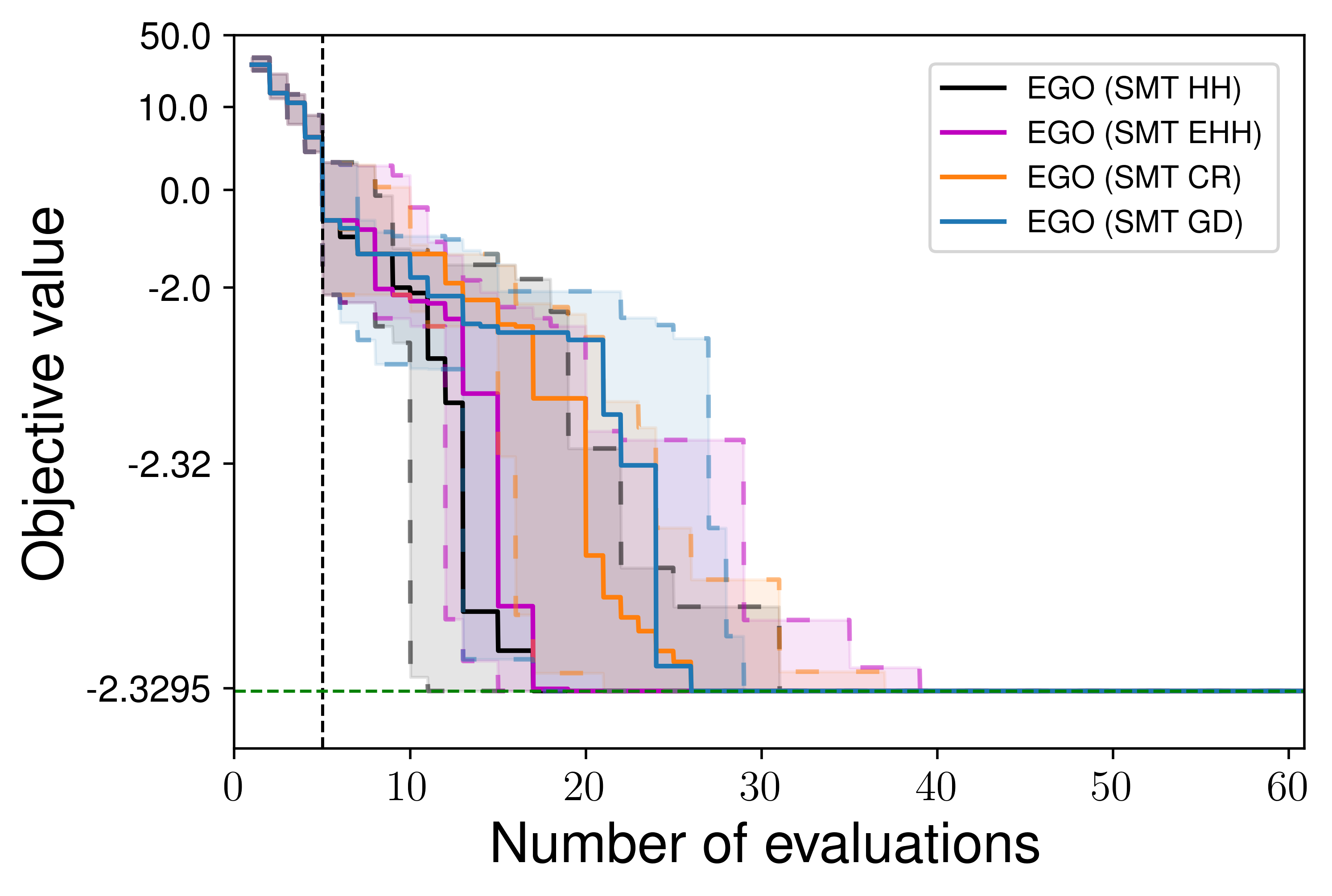}
\label{convmi}
}
\end{minipage}
\begin{minipage}[b]{.4\linewidth}
\centering 
\hspace{-1.5cm}
\subfloat[Boxplots after 20 evaluations.]{
\includegraphics[height=5cm,width=6.2cm]{  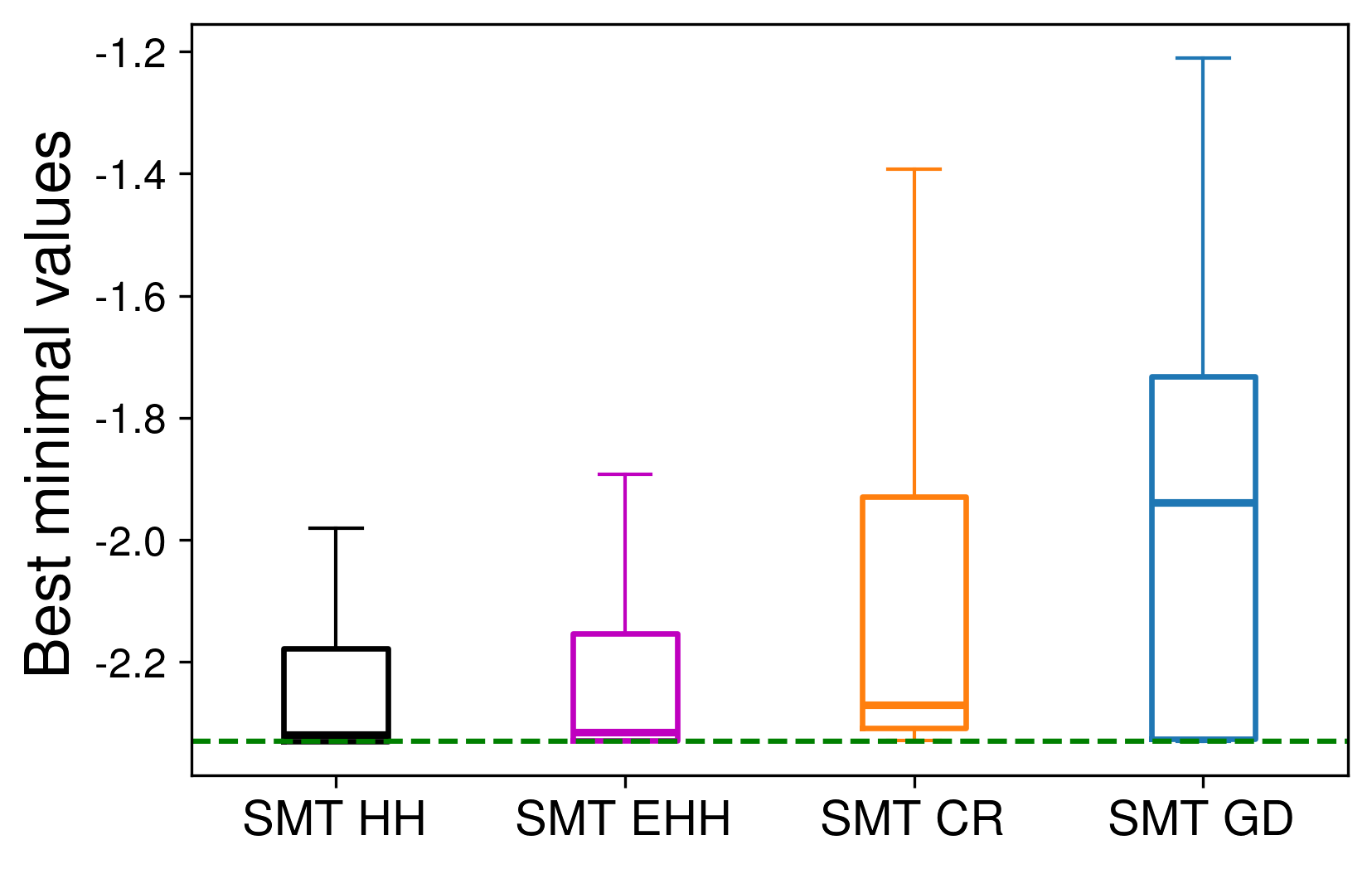}
\label{mini_mi}
}
\end{minipage}
\caption{Optimization results for the Toy function~\cite{CAT-EGO}.}
\label{res_optim_mi}
\end{figure}

In~\figref{fig:Branin_mixed} we illustrate how to use EGO with mixed variables based on the implementation of \texttt{SMT 2.0}. 
The illustrated problem is a mixed variant of the Branin function~\cite{AMIEGO}.

\begin{figure}[t!]
\begin{lstlisting}
#Import the Mixed Integer API
from smt.surrogate_models import KRG,MixIntKernelType,\
  DesignSpace,FloatVariable,IntegerVariable
from smt.applications.mixed_integer import \
  MixedIntegerSamplingMethod as misamp
#Define the function
from smt.problems import Branin
fun = Branin(ndim=2)
#Define the mixed design space
design_space = DesignSpace([
  IntegerVariable(*fun.xlimits[0]),
  FloatVariable(*fun.xlimits[1]),
])
#Perform a mixed integer sampling with LHS
from smt.sampling_methods import LHS
smp = misamp(LHS,design_space,random_state=42)
xdoe = smp(10)
# Call the Bayesian optimizer
from smt.applications import EGO
criterion = "EI"  #'EI' or 'SBO' or 'LCB'
ego = EGO(xdoe=xdoe,
  n_iter=20,
  criterion="EI",
  random_state=42,
  surrogate=KRG(design_space=design_space,
   categorical_kernel=MixIntKernelType.GOWER))
x_opt, y_opt, _, _, _ = ego.optimize(fun=fun)
# Check if the result is correct
self.assertAlmostEqual(0.494, float(y_opt), delta=1)
\end{lstlisting}

\caption{Example of usage of mixed surrogates for Bayesian optimization\textcolor{black}{.}}
\label{fig:Branin_mixed}
\end{figure}

Note that a dedicated notebook is available to reproduce the results presented in this paper and the mixed integer notebook also includes an extra mechanical application with composite materials~\cite{RaulAIAA}~\footnote{\url{https://colab.research.google.com/github/SMTorg/smt/blob/master/tutorial/SMT_MixedInteger_application.ipynb} }. 

\subsection{A hierarchical optimization problem} 
\label{sec:HV-BO}
The hierarchical test case considered in this paper to illustrate Bayesian  optimization is a modified Goldstein function~\cite{pelamattihier} detailed in~\ref{app:Goldstein}.
The resulting optimization problem involves 11 variables: 5 are continuous, 4 are integer (ordinal) and 2 are categorical.
These variables consist in 6 neutral variables, 1 dimensional (or meta) variable and 4 decreed variables.
Depending on the meta variable values, 4 different sub-problems can be identified.
The optimization problem is given by:
\begin{equation}
\begin{split}
& \min  f( x^{\cat}_{\neutral}, x^{\quant}_{\neutral}, x^{\cat}_{m}, x^{\quant}_{\decreed}  ) \\
& \mbox{w.r.t.} \ \  
x^{\cat}_{\neutral} =w_2 \in \{ 0,1 \} \\
& \quad \quad \quad x^{\quant}_{\neutral} = (x_1,x_2,x_5,z_3,z_4) \in \{ 0,100 \}^3 \times \{ 0,1,2 \}^2   \\
& \quad \quad \quad  x^{\cat}_{m} = w_1 \in \{ 0,1,2,3 \} \\
& \quad \quad \quad x^{\quant}_{\decreed} = (x_3,x_4,z_1,z_2) \in \{ 0,100 \}^2 \times \{ 0,1,2 \}^2   
\end{split}
\end{equation}
Compared to the model choice of~\citet{pelamattihier}, we chose to model $x_5$ and $w_2$ as neutral variables even if $f$ does not depend on $x_5$ when $w_2=0$. 
Other modeling choices are kept; for example, $w_2$ is a so-called "binary variable" and not a categorical one~\cite{muller_so-mi_2013}. 
Similarly, we also keep the formulation of $w_1$ as a categorical variable but a better model would be to model it as a "cyclic variable"~\cite{tran:hal-03170761}.
The resulting problem is described in~\ref{app:Goldstein}.
To assess the performance of our algorithm, we performed 20 runs with different initial DoE sampled by LHS.
Every DoE consists of $n+1=12$ points and we chose a budget of $5n = 55$ infill points.
To compare our method with a baseline, we also tested the random search method thanks to the \texttt{expand\_lhs} new method~\cite{condearenzana} described in Section~\ref{sec:sampling} and we also optimized the Goldstein function using EGO with a classic Kriging model based on imputation method (\texttt{Imp-Kernel}). 
This method replaces the decreed-excluded variables by their mean values: $50$ or $1$ respectively for $(x_3,x_4)$ and $(z_1,z_2)$. 
\figref{convhier} plots the convergence curves for the four methods. 
In particular, the median is the solid line and the first and third quantiles are plotted in dotted lines. 
To visualize better the corresponding data dispersion, the boxplots of the 20 best solutions are plotted in~\figref{mini_hier}. 
The results in~\figref{res_optim} show that the hierarchical Kriging models of \texttt{SMT 2.0} lead to better results than the imputation method or the random search both in terms of final objective value and variance over the 20 runs and in term of convergence rate.
More precisely, \texttt{SMT Arc-Kernel} and \texttt{SMT Alg-Kernel} Kriging model gave the best EGO results and managed to converge correctly as shown in~\figref{mini_hier}. 
More precisely, the 20 sampled DoEs led to similar performance and from one DoE, the method \texttt{SMT Alg-Kernel} managed to find the true minimum. However, this result has not been reproduced in other runs and is therefore not statistically significant. The variance between the runs is of similar magnitude regardless of the considered methods. 

\begin{figure}[H]
\begin{minipage}[b]{.6\linewidth}
\centering
\hspace{-1.25cm}
\subfloat[Convergence curves: medians of 20 runs.]{
\includegraphics[height=5cm,,width=7cm]
{  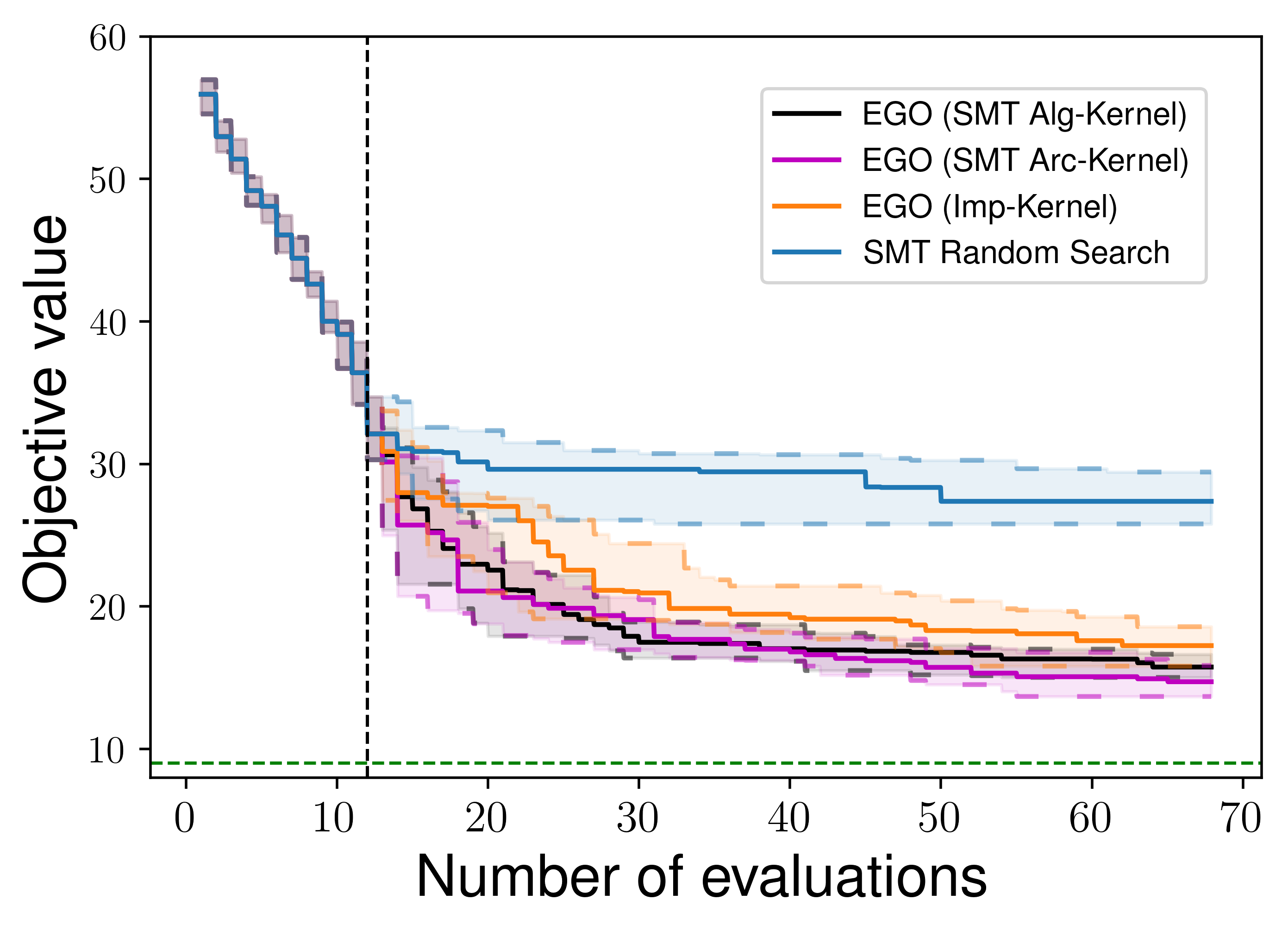}
\label{convhier}
}
\end{minipage}
\begin{minipage}[b]{.4\linewidth}
\centering 
\hspace{-1.5cm}
\subfloat[Boxplots after 67 iterations.]{
\includegraphics[height=5cm,width=6.2cm]{  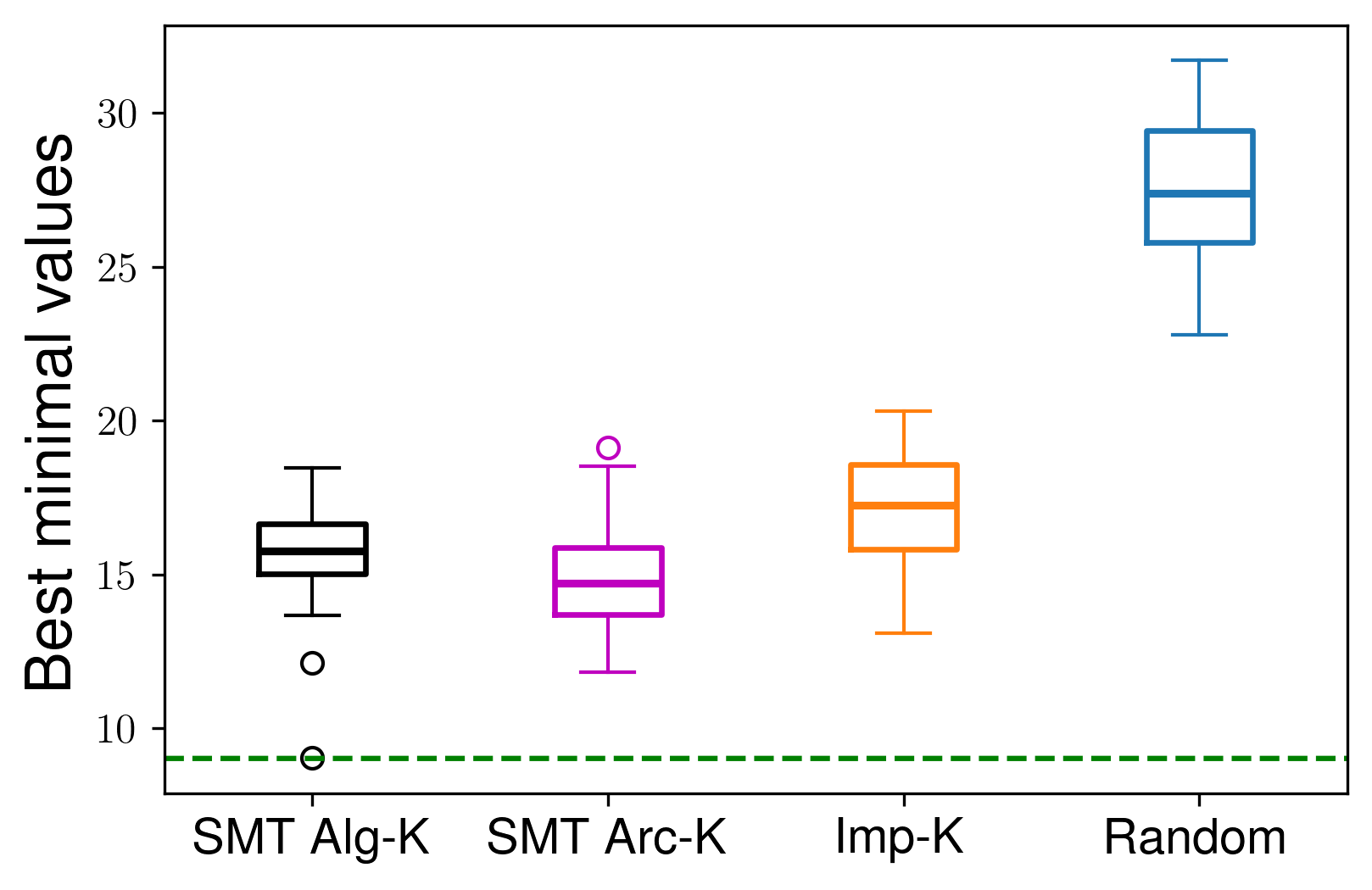}
\label{mini_hier}
}
\end{minipage}
\caption{Optimization results for the hierarchical Goldstein function.}
\label{res_optim}
\end{figure}

\section{Other relevant contributions in \texttt{SMT 2.0}}
\label{sec:other}
The new release \texttt{SMT 2.0} introduces several improvements besides Kriging for hierarchical and mixed variables. 
This section details the most important new contributions.  
Recall from Section~\ref{sec:organization} that five sub-modules are present in the code: \texttt{Sampling}, \texttt{Problems}, \texttt{Surrogate Models}, \texttt{Applications} and \texttt{Notebooks}.

\subsection{Contributions to \texttt{Sampling}}
\label{sec:sampling}
\paragraph{Pseudo-random Sampling}
The Latin Hypercube Sampling (LHS) is a stochastic sampling technique to generate quasi-random sampling distributions. 
It is among the most popular sampling method in computer experiments thanks to its simplicity and projection properties with high-dimensional problems. 
The LHS method uses the pyDOE package (Design Of Experiments for Python). 
Five criteria for the construction of LHS are implemented in SMT. 
The first four criteria (\texttt{center, maximin, centermaximin, correlation}) are the same as in pyDOE~\footnote{\url{https://pythonhosted.org/pyDOE/index.html}}.
The last criterion \texttt{ese}, is implemented by the authors of SMT~\cite{LHS}. 
In \texttt{SMT 2.0} a new LHS method was developed for the Nested design of experiments (\texttt{NestedLHS})~\cite{meliani2019multi} to use with multi-fidelity surrogates.
A new mathematical method (\texttt{expand\_lhs})~\cite{condearenzana} was developed in \texttt{SMT 2.0} to increase the size of a design of experiments while maintaining the \texttt{ese} property.
Moreover, we proposed a sampling method for mixed variables, and the aforementioned LHS method was applied to hierarchical variables in~\figref{res_optim}.

\subsection{Contributions to \texttt{Surrogate models}}
\label{sec:surrogates}
\paragraph{New kernels and their derivatives for Kriging}
Kriging surrogates are based on hyperparameters and on a correlation kernel.
Four correlation kernels are now implemented in \texttt{SMT 2.0}~\cite{Lee2011}.
In SMT, these correlation functions are absolute exponential (\texttt{abs\_exp}), Gaussian (\texttt{squar\_exp}), Matern 5/2 (\texttt{matern52}) and Matern 3/2 (\texttt{matern32}).
In addition, the implementation of gradient and Hessian for each kernel makes it possible to calculate both the first and second derivatives of the GP likelihood with respect to the hyperparameters~\cite{SMT2019}. 

\paragraph{Variance derivatives for Kriging}
To perform uncertainty quantification for system analysis purposes, it could be interesting to know more about the variance derivatives of a model~\cite{lopez, berthelin2022disciplinary, aerobest23cardoso}. 
For that purpose and also to pursue the original publication about derivatives~\cite{SMT2019}, \texttt{SMT 2.0} extends the derivative support to Kriging variances and kernels.

\paragraph{Noisy Kriging}
In engineering and in big data contexts with real experiments, surrogate models for noisy data are of significant interest.
In particular, there is a growing need for techniques like noisy Kriging and noisy Multi-Fidelity Kriging (MFK) for data fusion~\cite{platt2022systematic}.
For that purpose, \texttt{SMT 2.0} has been designed to accommodate Kriging and MFK to noisy data including the option to incorporate heteroscedastic noise (using the \texttt{use\_het\_noise} option) and to account for different noise levels for each data source~\cite{condearenzana}.

\paragraph{Kriging with partial least squares}
Beside MGP, for high-dimensional problems, the toolbox implements Kriging with partial least squares (KPLS)~\cite{bouhlel_KPLSK} and its extension KPLSK~\cite{Bouhlel18}. 
The PLS information is computed by projecting the data into a smaller space spanned by the principal components. 
By integrating this PLS information into the Kriging correlation matrix, the number of inputs can be scaled down, thereby reducing the number of hyperparameters required. 
The resulting number of hyperparameters $d_e$ is indeed much smaller than the original problem dimension $d$.
Recently, in \texttt{SMT 2.0}, we extended the KPLS method for multi-fidelity Kriging (MFKPLS and MFKPLSK)~\textcolor{black}{\cite{meliani2019multi, MFKPLS, charayron2023towards}}. 
We also proposed an automatic criterion to choose automatically the reduced dimension $d_e$ based on Wold's R criterion~\cite{wold_1975}.
This criterion has been applied to aircraft optimization with EGO where the number $d_e$ ($\texttt{n\_comp}$ in the code) for the model is automatically selected at every iteration~\cite{SciTech_cat}. 
Special efforts have been made to accommodate KPLS for multi-fidelity and mixed integer data~\textcolor{black}{\cite{MFKPLS,charayron2023towards}}.

\paragraph{Marginal Gaussian process} 
\texttt{SMT 2.0} implements Marginal Gaussian Process (MGP) surrogate models for high dimensional problems~\cite{EGORSE}. 
MGP are Gaussian processes taking into account hyperparameters uncertainty defined as a density probability function. 
Especially we suppose that the function to model $f : \Omega \mapsto \mathbb{R}$, where  $\Omega \subset \mathbb{R}^d$ and $d$ is the number of design variables, lies in a linear embedding $\mathcal{A}$ such as $\mathcal{A} = \{ u = Ax, x\in\Omega\},A \in \mathbb{R}^{d \times d_e}$ and $f(x)=f_{\mathcal{A}}(Ax)$ with $f(x)=f_{\mathcal{A}} : \mathcal{A} \mapsto \mathbb{R}$ and $d_e \ll d$.
Then, we must use a kernel $k(x,x')=k_{\mathcal{A}}(Ax,Ax')$ whose each component of the transfer matrix $A$ is an hyperparameter. 
Thus we have $d_e \times d$ hyperparameters to find. 
Note that $d_e$ is defined as $\texttt{n\_comp}$ in the code~\cite{MGP}. 

\paragraph{Gradient-enhanced neural network}
The new release \texttt{SMT 2.0} implements Gradient-Enhanced Neural Network (GENN) models~\cite{bouhlel2020}.
Gradient-Enhanced Neural Networks (GENN) are fully connected multi-layer perceptrons whose training process was modified to account for gradient information.
Specifically, the model is trained to minimize not only the prediction error of the response but also the prediction error of the partial derivatives: the chief benefit of gradient enhancement is better accuracy with fewer training points. 
Note that GENN applies to regression (single-output or multi-output), but not classification since there is no gradient in that case. 
The implementation is fully vectorized and uses ADAM optimization, mini-batch, and L2-norm regularization.
For example, GENN can be used to learn airfoil geometries from a database. 
This usage is documented in \texttt{SMT 2.0}~\footnote{\url{https://smt.readthedocs.io/en/latest/_src_docs/examples/airfoil_parameters/learning_airfoil_parameters.html}}.

\subsection{Contributions to \texttt{Applications}}
\paragraph{Kriging trajectory and sampling}
Sampling a GP with high resolution is usually expensive due to the large dimension of the associated covariance matrix.
Several methods are proposed to draw samples of a GP on a given set of points.
To sample a conditioned GP, the classic method  consists in using a Cholesky decomposition (or eigendecomposition) of the conditioned covariance matrix of the process but some numerical computational errors can lead to non SPD matrix.
A more recent approach  based on Karhunen-Loève decomposition of the covariance kernel with the Nyström method has been proposed in~\cite{betz2014numerical} where the paths can be sampled by generating independent standard Normal distributed samples. 
The different methods are documented in the tutorial \emph{Gaussian Process Trajectory Sampling}~\cite{menz2021variance}.  

\paragraph{Parallel Bayesian optimization}
Due to the recent progress made in hardware configurations, it has been of high interest to perform parallel optimizations.
A parallel criterion called qEI~\cite{Ginsbourger2010} was developed to perform Efficient Global Optimization (EGO): the goal is to be able to run batch optimization. 
At each iteration of the algorithm, multiple new sampling points are extracted from the known ones.
These new sampling points are then evaluated using a parallel computing environment.
Five criteria are implemented in \texttt{SMT 2.0}: Kriging Believer ({\footnotesize  \texttt{KB}}), Kriging Believer Upper Bound ({\footnotesize \texttt{KBUB}}), Kriging Believer Lower Bound ({\footnotesize \texttt{KBLB}}), Kriging Believer Random Bound ({\footnotesize \texttt{KBRand}}) and Constant Liar ({\footnotesize \texttt{CLmin}})~\cite{roux2020efficient}.

\section{\textcolor{black}{Conclusion}}
\label{sec:concl}

\texttt{SMT 2.0} introduces significant upgrades to the Surrogate Modeling Toolbox.
This new release adds support for hierarchical and mixed variables and improves the surrogate models with a particular focus on Kriging (Gaussian process) models. 
\texttt{SMT 2.0} is distributed through an open-source license and is freely available online~\footnote{ \url{https://github.com/SMTorg/SMT}}.
We provide documentation that caters to both users and potential developers~\footnote{\url{https://smt.readthedocs.io/en/latest/}}. 
\texttt{SMT 2.0} enables users and developers collaborating on the same project to have a common surrogate modeling tool that facilitates the exchange of methods and reproducibility of results.

SMT has been widely used in aerospace and mechanical modeling applications. 
Moreover, the toolbox is general and can be useful for anyone who needs to use or develop surrogate modeling techniques, regardless of the targeted applications. 
SMT is currently the only open-source toolbox that can build hierarchical and mixed surrogate models.

\section*{\footnotesize Data availability}
\vspace{-0.1cm}
\noindent \footnotesize  Data will be made available on request. Results can be reproduced freely online at  \newline \noindent \scriptsize \url{https://colab.research.google.com/github/SMTorg/smt/blob/master/tutorial/NotebookRunTestCases_Paper_SMT_v2.ipynb}.

\vspace{-0.2cm}
\section*{\footnotesize Supplementary material}
\vspace{-0.1cm}
\noindent  \footnotesize  Supplementary material related to this article can be found online at  \url{https://doi.org/10.1016/j.advengsoft.2023.103571}.
\section*{Acknowledgements}


We want to thank all those who contribute to this release.
Namely,
M. A. Bouhlel, 
I. Cardoso,
R. Carreira Rufato,
R. Charayron,
R. Conde Arenzana,
S. Dubreuil,
A. F. López-Lopera,
M. Meliani,
M. Menz,
N. Moëllo,
A. Thouvenot,
R. Priem,
E. Roux 
and
F. Vergnes.
This work is part of the activities of ONERA - ISAE - ENAC joint research group. 
We also acknowledge the partners institutions: ONERA, NASA Glenn, ISAE-SUPAERO, Institut Clément Ader (ICA), the University of Michigan, Polytechnique Montréal and the University of California San Diego. 

The research presented in this paper has been performed in the framework of the AGILE 4.0 project (Towards cyber-physical collaborative aircraft development), funded by the European Union Horizon 2020 research and innovation framework programme under grant agreement n${\mbox{}^\circ}$ {815122} and in the COLOSSUS project (Collaborative System of Systems Exploration of Aviation Products, Services and Business Models) funded by the European Union Horizon Europe research and innovation framework programme under grant agreement n${\mbox{}^\circ}$ {101097120}.

We also are grateful to E. Hallé‑Hannan from Polytechnique Montréal for the hierarchical variables framework.

\appendix


\section{Toy test function}
\label{app:Toy}
This Appendix gives the detail of the toy function of Section~\ref{sec:MI-BO}~\footnote{\url{https://github.com/jbussemaker/SBArchOpt}}.
First, we recall the optimization problem:
\begin{equation}
\begin{split}
& \min  f( x^{\cat}, x^{\quant}) \\
& \mbox{w.r.t.} \ \  x^{\cat} = c_1 \in \{ 0,1,2,3,4,5,6,7,8,9 \} \\
& \quad \quad \quad x^{\quant} = x_1 \in [ 0,1 ] \\
\end{split}
\end{equation}
The toy function $f$ is defined as
\begin{equation}
\begin{split}
   f({x}, {c_1})  =&   \mathds{1}_{c_1=0}  \ \cos(3.6 \pi(x-2)) +x -1 \\
     + &\mathds{1}_{c_1=1} \ 2 \cos(1.1 \pi \exp(x)) - \frac{x}{2} +2  \\
     + &\mathds{1}_{c_1=2}  \  \cos ( 2 \pi x)  + \frac{1}{2}x \\
  +&\mathds{1}_{c_1=3}  \  x ( \cos(3.4 \pi (x-1)) - \frac{x-1}{2})\\
  +& \mathds{1}_{c_1=4}  \   - \frac{x^2}{2} \\
 + &\mathds{1}_{c_1=5}  \  2 \cos(0.25 \pi \exp( -x^4))^2 - \frac{x}{2} +1 \\ 
 +& \mathds{1}_{c_1=6}   \ x \cos(3.4 \pi x ) - \frac{x}{2} +1 \\ 
 +&\mathds{1}_{c_1=7}   \  - x  (\cos(3.5 \pi x ) + \frac{x}{2}) +2 \\ 
 + &\mathds{1}_{c_1=8}   \ - \frac{x^5}{2} +1 \\ 
 +& \mathds{1}_{c_1=9}  \  - \cos (2.5 \pi x)^2 \sqrt{x} - 0.5 \ln (x+0.5)  - 1.3\\ 
\end{split}
\end{equation}

\section{Hierarchical Goldstein test function}
\label{app:Goldstein}
This Appendix gives the detail of the hierarchical Goldstein problem of Section~\ref{sec:HV-BO}~\footnote{\url{https://github.com/jbussemaker/SBArchOpt}}.
First, we recall the optimization problem:
\begin{equation}
\begin{split}
& \min  f( x^{\cat}_{\neutral}, x^{\quant}_{\neutral}, x^{\cat}_{m}, x^{\quant}_{\decreed}  ) \\
& \mbox{w.r.t.} \ \
x^{\cat}_{\neutral} =w_2 \in \{ 0,1 \} \\
& \quad \quad \quad x^{\quant}_{\neutral} = (x_1,x_2,x_5,z_3,z_4) \in [ 0,100 ] ^3 \times \{ 0,1,2 \}^2  \\
& \quad \quad \quad   x^{\cat}_{m} = w_1 \in \{ 0,1,2,3 \} \\
& \quad \quad \quad 
 x^{\quant}_{\decreed} = (x_3,x_4,z_1,z_2) \in [ 0,100 ] ^2 \times \{ 0,1,2 \}^2   
\end{split}
\end{equation}
The hierarchical and mixed function $f$ is defined as a hierarchical function that depends on $f_0$,  $f_1$, $f_2$ and $Gold_\text{cont}$ as describes in the following.
\begin{equation}
\begin{split}
   &f({x_1}, {x_2}, {x_3}, {x_4}, {z_1}, {z_2}, {z_3}, {z_4}, {x_5}, {w_1}, {w_2})  = \\
    &  \quad \  \mathds{1}_{w_1=0} f_0( {x_1}, {x_2},  {z_1}, {z_2}, {z_3}, {z_4}, {x_5}, {w_2}) \\
    & + \mathds{1}_{w_1=1} f_1 ({x_1}, {x_2}, {x_3}, {z_2}, {z_3}, {z_4}, {x_5}, {w_2})  \\
    & + \mathds{1}_{w_1=2} f_2 ({x_1}, {x_2}, {x_4}, {z_1}, {z_3}, {z_4}, {x_5}, {w_2})  \\
   & + \mathds{1}_{w_1=3}  Gold_{\text{cont}} ({x_1}, {x_2}, {x_3}, {x_4}, {z_3}, {z_4}, {x_5},  {w_2}).
\end{split}
\end{equation}
Then, the functions $f_0$, $f_1$ and $f_2$ are defined as mixed variants of $Gold_\text{cont}$ as such
\begin{equation}
\begin{split}
   &f_0( {x_1}, {x_2},  {z_1}, {z_2}, {z_3}, {z_4}, {x_5}, {w_2})  = \\
       & \mathds{1}_{z_2=0} \big(  \mathds{1}_{z_1=0}  Gold_{\text{cont}} ({x_1}, {x_2}, 20,20, {z_3}, {z_4}, {x_5},  {w_2})  \\
      & \quad   + \mathds{1}_{z_1=1} Gold_{\text{cont}} ({x_1}, {x_2}, 50,20, {z_3}, {z_4}, {x_5},  {w_2})   \\
      & \quad  + \mathds{1}_{z_1=2}Gold_{\text{cont}} ({x_1}, {x_2}, 80,20, {z_3}, {z_4}, {x_5},  {w_2})   \big) \\
       & \mathds{1}_{z_2=1} \big(  \mathds{1}_{z_1=0}  Gold_{\text{cont}} ({x_1}, {x_2}, 20,50, {z_3}, {z_4}, {x_5},  {w_2})  \\
      & \quad   + \mathds{1}_{z_1=1} Gold_{\text{cont}} ({x_1}, {x_2}, 50,50, {z_3}, {z_4}, {x_5},  {w_2})   \\
      & \quad  + \mathds{1}_{z_1=2}Gold_{\text{cont}} ({x_1}, {x_2}, 80,50, {z_3}, {z_4}, {x_5},  {w_2})   \big) \\
      & \mathds{1}_{z_2=2} \big(  \mathds{1}_{z_1=0}  Gold_{\text{cont}} ({x_1}, {x_2}, 20,80, {z_3}, {z_4}, {x_5},  {w_2})  \\
      & \quad   + \mathds{1}_{z_1=1} Gold_{\text{cont}} ({x_1}, {x_2}, 50,80, {z_3}, {z_4}, {x_5},  {w_2})   \\
      & \quad  + \mathds{1}_{z_1=2}Gold_{\text{cont}} ({x_1}, {x_2}, 80,80, {z_3}, {z_4}, {x_5},  {w_2})   \big) \\
  \end{split}
  \end{equation}
  \begin{equation*}
  \begin{split}
   &f_1 ({x_1}, {x_2}, {x_3}, {z_2}, {z_3}, {z_4}, {x_5}, {w_2})  = \\
       &  \mathds{1}_{z_2=0}  Gold_{\text{cont}} ({x_1}, {x_2}, {x_3},20, {z_3}, {z_4}, {x_5},  {w_2})  \\
      & \quad   + \mathds{1}_{z_2=1} Gold_{\text{cont}} ({x_1}, {x_2}, {x_3},50, {z_3}, {z_4}, {x_5},  {w_2})   \\
      & \quad  + \mathds{1}_{z_2=2}Gold_{\text{cont}} ({x_1}, {x_2}, {x_3},80, {z_3}, {z_4}, {x_5},  {w_2})  \\[12pt]
   &f_2 ({x_1}, {x_2}, {x_4}, {z_1}, {z_3}, {z_4}, {x_5}, {w_2})  = \\
       &  \mathds{1}_{z_1=0}  Gold_{\text{cont}} ({x_1}, {x_2}, 20, {x_4}, {z_3}, {z_4}, {x_5},  {w_2})  \\
      & \quad   + \mathds{1}_{z_1=1} Gold_{\text{cont}} ({x_1}, 50, {x_2}, {x_4}, {z_3}, {z_4}, {x_5},  {w_2})   \\
      & \quad  + \mathds{1}_{z_1=2}Gold_{\text{cont}} ({x_1}, {x_2}, 80, {x_4}, {z_3}, {z_4}, {x_5},  {w_2})  
\end{split}
\end{equation*}
To finish with, the function $Gold_{\text{cont}}$ is given by
\begin{equation}
\begin{split}
    & Gold_{\text{cont}} ({x_1}, {x_2}, {x_3}, {x_4}, {z_3}, {z_4}, {x_5},  {w_2}) = 
        53.3108
        + 0.184901   {x_1}  \\
        &- 5.02914   {x_1} ^3   .10    ^{-6}
        + 7.72522    {x_1} ^{z_3}    .10    ^{-8}
        - 0.0870775    {x_2}
        - 0.106959    {x_3}  \\
        &+ 7.98772    {x_3} ^{z_4}    .10    ^{-6} 
         + 0.00242482    {x_4}
        + 1.32851    {x_4} ^3    .10    ^{-6}
        - 0.00146393    {x_1}    {x_2}\\
        &- 0.00301588    {x_1}    {x_3} 
         - 0.00272291    {x_1}    {x_4}
        + 0.0017004    {x_2}    {x_3}
        + 0.0038428    {x_2}    {x_4}\\
       & - 0.000198969    {x_3}    {x_4} 
        + 1.86025    {x_1}    {x_2}    {x_3}    .10    ^{-5}
        - 1.88719    {x_1}    {x_2}    {x_4}    .10    ^{-6}\\
       & + 2.50923    {x_1}    {x_3}    {x_4}    .10    ^{-5} 
        - 5.62199    {x_2}    {x_3}    {x_4}    .10    ^{-5} 
        +  {w_2} \left( 5    \cos \left( \frac{ 2 \pi}{100}  x_5 \right) - 2\right).
\end{split}
\end{equation}

\bibliographystyle{abbrvnat}
\setlength{\bibsep}{-1pt}
\bibliography{main.bib}
\end{document}